\documentclass[conference]{IEEEtran}
\IEEEoverridecommandlockouts% This command is only needed if 
\usepackage{times}

\makeatletter 
\let\NAT@parse\undefined
\makeatother

\usepackage{multicol}
\usepackage[bookmarks=true]{hyperref}

\usepackage{bbm}
\usepackage{calc}
\usepackage{url}
\usepackage{transparent}
\usepackage{hyperref}
\hypersetup{
  colorlinks =false,
  urlcolor = black,
  linkcolor = black
}
\usepackage{graphicx}
\usepackage[cmex10]{amsmath}
\usepackage{bm}
\usepackage{amssymb}
\usepackage{rotating}

\usepackage{chngcntr}
\counterwithin{paragraph}{subsection} % makes paragraph depend on subsection

\usepackage{nicefrac}
\usepackage{cite}
\usepackage[caption=false,font=footnotesize]{subfig}
\usepackage[usenames, dvipsnames]{color}
\usepackage{colortbl}
\usepackage{overpic}
\graphicspath{{./},{./pictures/pdf/},{./pictures/ps/},{./pictures/png/},{./pictures/jpg/}}
\usepackage{breqn} %for breaking equations automatically
\usepackage[ruled]{algorithm}
\usepackage{algpseudocode}
\usepackage{multirow}
\usepackage{todonotes}
\usepackage{authblk}
\usepackage{balance}

\providecommand{\norm}[1]{\left\lVert#1\right\rVert}

\newcommand{\figwid}{0.22\columnwidth}

\allowdisplaybreaks

\title{\LARGE \bf Steering a Particle Swarm Using Global Inputs\\ and Swarm Statistics}

\author{Shiva Shahrokhi, Lillian Lin, Chris Ertel, Mable Wan,  Aaron T. Becker% end author block 
\thanks{This work was supported by the National Science Foundation under Grant No.\ \href{http://nsf.gov/awardsearch/showAward?AWD_ID=1553063}{[IIS-1553063]}.}% <-this % stops a space
\thanks{Authors are with the Department of Electrical and Computer Engineering,  University of Houston, Houston, TX 77204 USA        {\tt\small  \{sshahrokhi2, atbecker\}@uh.edu}}
}

\begin{document}

\maketitle
\thispagestyle{empty}
\pagestyle{empty}
\begin{abstract}
Microrobotics has the potential to revolutionize many applications including targeted material delivery, assembly, and surgery.  The same properties that promise breakthrough solutions---small size and large populations---present unique challenges for controlling motion. 

Robotic manipulation usually assumes intelligent agents, not particle systems manipulated by a global signal.
To identify the key parameters for particle manipulation, we used a collection of online games where players steer swarms of up to 500 particles to complete manipulation challenges. We recorded statistics from over ten thousand players. 
Inspired by techniques where human operators performed well, we investigate controllers that use only the mean and variance of the swarm. We prove the mean position is controllable and provide conditions under which variance is controllable.  We next derive automatic controllers for these and a hysteresis-based switching control to regulate the first two moments of the particle distribution. 
Finally, we employ these controllers as primitives for an object manipulation task and implement all controllers on 100 kilobots controlled by the direction of a global light source.
%This paper examines object manipulation by a swarm of particles, each actuated by the same shared global input.
%Microrobotics has the potential to revolutionize many applications including targeted material delivery, assembly, and surgery.  The same properties that promise breakthrough solutions---small size and large populations---present unique challenges for controlling motion. 
%
%Robotic manipulation usually assumes intelligent agents, not particle systems manipulated by a global signal.
%To identify the key parameters for particle manipulation, we used
%a collection of online games where players steer swarms of up to 500 particles to complete manipulation challenges. We recorded statistics from over ten thousand players. 
% Inspired by techniques where human operators performed well, this paper next investigates controllers that use only the mean and variance of the swarm. We prove that the mean position is controllable and then provide conditions under which variance is controllable.  We next derive automatic controllers for these and a hysteresis-based switching control to regulate the first two moments of the particle distribution. 
% Finally, we employ these controllers as primitives for an object manipulation task and implement all the automatic controllers on 100 kilobots controlled by the direction of a global light source.
% This automatic controller uses policy iteration for path planning, handles outliers by partitioning the workspace, and minimizes pushing the object backwards using potential field navigation.
\end{abstract}

%IJRR \keywords{ global control, swarm, human-swarm interaction, micro-robot}
%  TRO       Manipulation Planning, 	Underactuated Robots,   human-swarm interaction,   swarm
%  maybe TRO	Manipulation and Compliant Assembly, 	Path Planning for Multiple Mobile Robot Systems

%Previously presented*  
% [1] A. T. Becker, C. Ertel, and J. McLurkin, ÒCrowdsourcing swarm manipulation experiments: A massive online user study with large swarms of simple robots,Ó in IEEE International Conference on Robotics and Automation (ICRA), 2014, pp. 2825Ð2830.
% [2] S. Shahrokhi and A. T. Becker, ÒStochastic swarm control with global inputs,Ó in IEEE/RSJ International Conference on  Intelligent Robots and Systems (IROS), Sep. 2015, pp. 421Ð427.

%\input{00-Abstract.tex}
\section{Introduction}\label{sec:Intro}
%This project studies system models and user interfaces for five multi-robot manipulation tasks with large populations of micro- and nanorobots.  We test several system models with different limitations on controllability and observability of the motion controller, and evaluate several different user interfaces.  We conduct user experiments to understand the impact of these limitations and design choices. 

%Micro- and nanorobotics have the potential to revolutionize many applications including targeted material delivery, assembly, and surgery.  The same properties that promise breakthrough solutions---small size and large populations---present unique challenges to generating controlled motion.  
Large populations of micro- and nanorobots are being produced in laboratories around the world, with diverse potential applications in drug delivery and construction, see \cite{Peyer2013,Shirai2005,Chiang2011}. These activities require robots that behave intelligently.
Limited computation and communication rules out autonomous operation or direct control over individual units; instead we must rely on global control signals broadcast to the entire particle population.    
This paper examines object manipulation by a swarm of particles, each actuated by the same shared global input, as illustrated in Fig.~{\ref{fig:bigPictureMeanAndVarianceForSwarm}. 
The transportation methodology is similar to that in \cite{sugawara2014object}, but rather than using onboard computation or sensing, the particles all move in the same direction.

%%%% Move it to the cover letter.
%This paper combines the content of two preliminary conference papers, extending their substance and providing full details in a single journal paper.
 % \cite{swarmcontrol2013} covered the first three months of SwarmControl.net experiments, and \cite{ShahrokhiIROS2015} presented simulations of object manipulation.  This paper presents three years of results from new games at SwarmControl.net.  For object manipulation, this paper presents robust new algorithms for manipulation, path planning, and obstacle avoidance, and a rich set of parametric studies.  All hardware validation experiments are new.

Many promising applications for particle swarms require direct human control, but user interfaces to thousands---or millions---of particles is a daunting human-swarm interaction (HSI) challenge. Our early work with over a hundred hardware robots and thousands of simulated particles demonstrated that direct human control of large swarms is possible, \cite{Becker2013b}. 
Unfortunately, the logistical challenges of repeated experiments with over one hundred robots prevented large-scale tests. 
There is currently no comprehensive understanding of user interfaces for controlling multi-robot systems with massive populations.  
One contribution of this paper is a tool for investigating HSI methods through statistically significant numbers of experiments.

\begin{figure}
\centering
\begin{overpic}[width=\columnwidth]{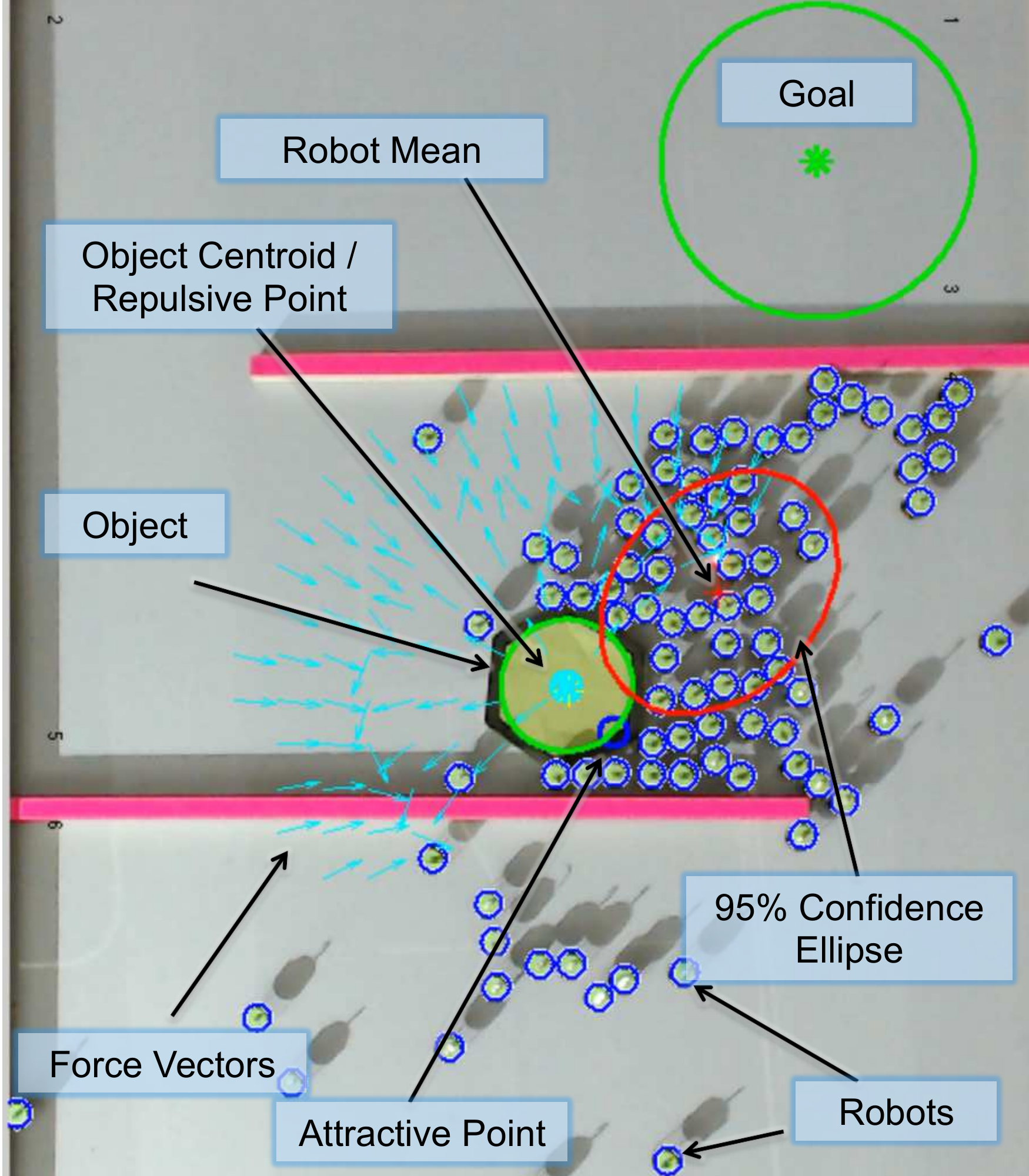}\end{overpic}
\caption{\label{fig:bigPictureMeanAndVarianceForSwarm} 
A swarm of particles, all actuated by a uniform control input where each particle gets the same control input, can be effectively manipulated by a control law that uses only the mean and variance of the robot distribution.  
Here a swarm of particles (kilobot robots) pushes a green hexagon toward the goal (see video attachment).
}
\end{figure}

%Inspired by our online experiments, because it is not always possible to gather pose information on each robot for feedback control and
%It is not always practical to gather pose information on individual robots for feedback control; the robots might be difficult or impossible to sense individually due to their size and location. However, it is often possible to sense global properties of the group, such as mean position and density.
Often particles are difficult or impossible to sense individually due to their size and location. 
For example, microrobots are smaller than the minimum resolution of a clinical MRI-scanner, see \cite{martel2014computer}, however it is often possible to sense global properties of the group such as mean position and variance. 
To make progress in automatic control with global inputs, this paper presents swarm manipulation controllers inspired by our online experiments that require only mean and variance measurements of the particle's positions. 
 To perform the object manipulation task illustrated in Fig.~\ref{fig:bigPictureMeanAndVarianceForSwarm}, we use these controllers   as primitives, policy iteration for path planning, handle outliers by partitioning the workspace, and minimize pushing the object backwards with potential field navigation.

 %function and implementation function

 Our paper is organized as follows.  After a discussion of related work in \S \ref{sec:RelatedWork},  we describe our experimental methods for an online human-user experiment and their results in \S \ref{sec:expMethods}. Next we prove that the mean and variance of a particle swarm are controllable in \S \ref{sec:theory}, and present automatic controllers in \S \ref{sec:simulation}. We use these controllers as primitives and present a framework for manipulating an object through a maze in \S \ref{sec:exp}. 
 We conclude with implementations of these controllers in our hardware robots and use them to complete an object manipulation task with 100 kilobots in \S \ref{sec:realExperiment}.

%%%%%%%%%%%%%%%%%%%%%%%%%%%%%%%%%%%%%%%%%%%%%%%%%%%%%%%%%%%
\section{Related Work}\label{sec:RelatedWork}
%%%%%%%%%%%%%%%%%%%%%%%%%%%%%%%%%%%%%%%%%%%%%%%%%%%%%%%%%%%
%Global-Control of Microrobots
%Block-Pushing and Compliant Manipulation
%Human-Swarm Interaction
%Microrobots can be produced in large numbers, but are generally controlled by global inputs.  This leads to challenges in sensing, 
This section describes global control challenges and reviews highlights of human-swarm interaction, block pushing, and compliant manipulation.

\subsection{Global control of microrobots}

This paper investigates global control of particles that have no onboard computation. This prevents us from applying controllers that require computation on the agents, as in \cite{milutinovic2006modeling,prorok2011multi,demir2015probabilistic}.
%\cite{milutinovic2006modeling} requires robots that can execute task primitives.  Our particles simply move in the direction of the global control unless they are stopped by an obstacle. 
%\cite{palmer2011hamiltonian} uses only Open-Loop control.  Our submission relies on state feedback 
%\cite{prorok2011multi} requires collective intelligence of a group of miniature robots
%\cite{demir2015probabilistic} requires robots that can individually decide when to actuate and when to remain anchored, can perform onboard computation (generate a random number, calculate its bin and the next bin). 
Another control paradigm is to construct robots with physical heterogeneity so that they respond differently to a global broadcast control signal.  Examples include \emph{scratch-drive microrobots}, actuated and controlled by a DC voltage signal from a substrate by \cite{Donald2006,Donald2008};  magnetic structures  with different cross-sections that can be independently steered by \cite{Floyd2011,Diller2013};  \emph{MagMite} microrobots with different resonant frequencies and a global magnetic field by \cite{Frutiger2008}; and  magnetically controlled nanoscale helical screws constructed to stop movement at different cutoff frequencies of a global magnetic field by
\cite{Tottori2012} and \cite{Peyer2013}. 
Similarly, our previous work, \cite{Becker2012,Becker2012k}, focused on exploiting inhomogeneity between robots.  These control algorithms theoretically apply to any number of robots, even robotic continuums.

However, process noise cancels the differentiating effects of inhomogeneity for more than tens of robots.  We desire control algorithms that extend to many thousands of robots.
Limited position control was achieved by \cite{bretl2007control} and our previous work \cite{beckerIJRR2014}, but both used robots commanded in their local coordinate frame. Our new submission focuses on a more common paradigm: particles commanded in a global coordinate frame

%  \subsection{Three challenges for massive manipulation}
 While it is now possible to create many microrobots, there remain challenges in control, sensing, and computation:
  
 \paragraph{Control---global inputs}
 Many micro- and nanorobotic systems, see \cite{Tottori2012,Shirai2005,Chiang2011,Donald2006,Donald2008,Takahashi2006,Floyd2011,Diller2013,Frutiger2008,Peyer2013}
   rely on global inputs, where each robot receives an exact copy of the control signal.  Our experiments follow this global model.
   %Two reasonable questions are ``What tasks are possible with many robots, all under uniform control inputs?'' and ``What tasks are impossible with many robots, all under uniform control inputs?'' 
  
 \paragraph{Sensing---large populations}
$n$ differential-drive robots in a 2D workspace require $3n$ state variables. Even holonomic robots require $2n$ state variables. Numerous methods exist for measuring this state in microrobotics \cite{Peyer2013,Chiang2011,martel2014computer}.  These solutions use computer vision systems to sense position and heading angle, with corresponding challenges of handling missed detections and image registration between detections and robots.  These challenges increase at small scales where sensing competes with control for communication bandwidth.   We examine control when the operator has access to partial feedback, including only the first and/or second moments of a population's position, or only the convex-hull containing the robots.
 
\paragraph{Computation---calculating the control law}
In our previous work the controllers required at best a summation over all the robot states, see \cite{Becker2012k} and at worst a matrix inversion, see \cite{Becker2012}. 
These operations become intractable for large populations of robots. By focusing on \emph{human} control of large robot populations, we accentuate computational difficulties because the controllers are implemented by the unaided human operator. % We present an approach that, for many tasks, bypasses these problems due to large populations by allowing the user to command the entire population as a single unit.  For position control we cannot bypass this problem, but we provide an algorithm that scales linearly in the number of robots. 

 \subsection{Human-swarm interaction}
 
  Most humans are able to, with practice, steer a swarm of robots controlled by a global input. Prior to our paper, no algorithm existed. Using human input to learn how to control a dynamic system is a line of research with a rich history \cite{argall2009, abbeel2010}. This paper exploits insights gained from \href{http://www.swarmcontrol.net}{\emph{SwarmControl.net}}, particularly the fact that having a swarm's mean and variance is sufficient for object manipulation through an obstacle field.
  
 A user interface enabling an operator to maneuver a swarm of robots through a cluttered workspace by specifying the bounding prism for the swarm and then translating or scaling this prism is designed in \cite{ayanian2014controlling}. Our paper shares the concept of a global control input, but our robots have no onboard computation and cannot track a virtual boundary.
 
Human \emph{fanout}, the number of robots a single human user could directly control is studied in \cite{Jr2004}.  %Studied as a key aspect of human-robot interaction, 
They postulated that the optimal number of robots was approximately  the autonomous time  divided by the interaction time required by each robot.  Their sample problem involved a multi-robot search task, where users could assign goals to robots.  Their user interaction studies with simulated planar robots  indicated a \emph{fanout plateau} of about 8 robots, with diminishing returns for more robots.   They hypothesized the location of this plateau is highly dependent on the underlying task.  Indeed, our paper indicates there are tasks without plateaus. % (Lemmings is an example of this as well, for systems with autonomy
Their research investigated robots with 3 levels of autonomy.  We use robots without autonomy, corresponding with their first-level robots.

% Removed this -- 
%Chen, Barnes, and Harper-Sciarini published a review of supervisory control but emphasized high levels of autonomy, using the 10-level taxonomy given by \cite{Parasuraman2000}.  The direct control techniques this paper examines are the lowest level.

%\cite{Squire:2006:HCM:1121241.1121248} designed experiments showing user-interface design had a high impact on the task effectiveness and the number of robots that could be controlled simultaneously in a multi-robot task.

Several user studies compare methods for controlling large swarms of simulated robots, for example \cite{bashyal2008human,de2012controllability,kolling2012towards}.  These studies provide insights but are limited by cost to small user studies; have a closed-source code base; and focus on controlling intelligent, programmable agents.  
For instance, the studies \cite{bashyal2008human},  \cite{de2012controllability}, and \cite{kolling2012towards}  were limited to a pool of 5, 18, and 32  participants.
	Using an online testing environment, we conduct similar studies but with sample sizes three orders of magnitude larger.
%Crowdsourcing human-swarm interaction is gaining considerable attention, see \cite{tavakoli2016crowdsourced} and \cite{hauert2014mechanisms}.

\subsection{Block pushing and compliant manipulation}
Unlike \emph{caging} manipulation, where robots form a rigid arrangement around an object, as in \cite{Sudsang2002,Fink2008}, our swarm of robots is unable to grasp the blocks they push, and so our manipulation strategies are similar to \emph{nonprehensile manipulation} techniques, e.g.~\cite{Lynch1999}, where forces must be applied along the center of mass of the moveable object. A key difference is that our robots are compliant and tend to flow around the object, making this similar to fluidic trapping as in \cite{Armani2006} and \cite{Becker2009}.  

Our $n$-robot system with 2 control inputs and 4$n$ states is inherently under-actuated, and superficially bears resemblance to compliant, under-actuated manipulators.  Our swarms conform to the object to be manipulated, but lack the restoring force provided by flexures in \cite{odhner2014compliant} or silicone in \cite{deimel2014novel}.  Our swarms tend to disperse and so to regroup them we require artificial forces like the variance control primitives in \S \ref{sec:VarianceControl}.

\subsection{Relationship to authors' prior work}
 This paper combines the content of two preliminary conference papers, extending their substance and providing full details in a single journal paper.  One paper covered the first three months of SwarmControl.net experiments~\cite{swarmcontrol2013}, and the second presented simulations of object manipulation~\cite{ShahrokhiIROS2015}. This paper presents three years of results from SwarmControl.net. For object manipulation, this paper presents robust new algorithms for manipulation, path planning, and obstacle avoidance, and a rich set of parameter sweeps over key variables. All hardware validation experiments are new.

\section{Online experiment}
\label{sec:expMethods}
%%%%%%%%%%%%%%%%%%%%%%%%%%%%%%%%%%%%%%%%%%%%%%%%%%%%%%%%%%%

% Experimental Methods
%%  Platform  
%%  Human Subjects
%%% recruited through social media
%%% IRB form  Protocol Number: 14-012E Protocol Title: Massive Manipulation: A n online user study on controlling large swarms of simple robot sApproval Date: 7/26/2013Expiration Date: 7/26/2014
%%% Costs  for experiment:  ??
%%% Instrumenting:
%%%% Google analytics, airbrake, etc.

%% wherein we describe our framework
\begin{figure}
%\centering
\renewcommand{\figwid}{0.3\columnwidth}
\begin{overpic}[width =\figwid]{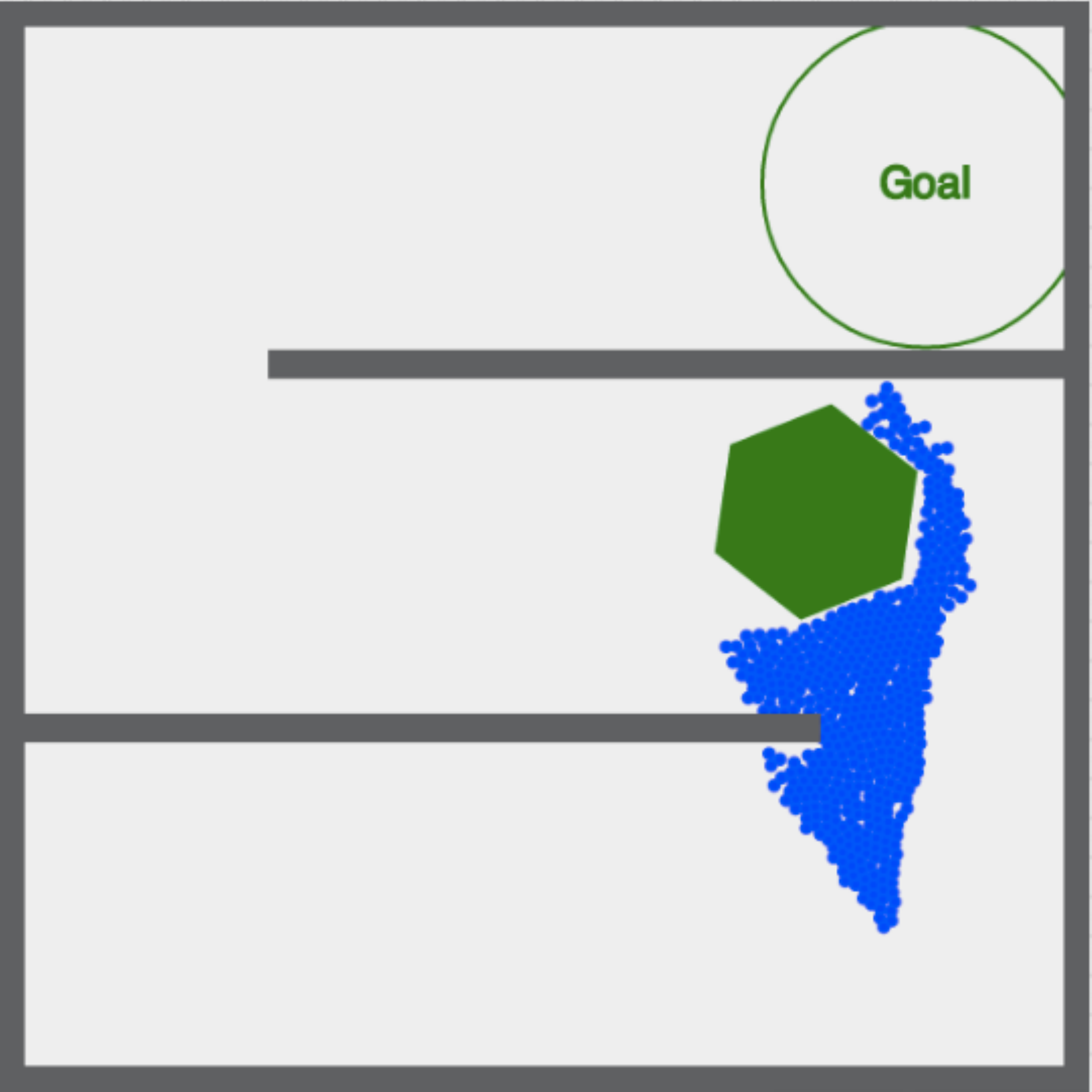}	\put(10,80){\textbf{a} }\end{overpic}~
\begin{overpic}[width =\figwid]{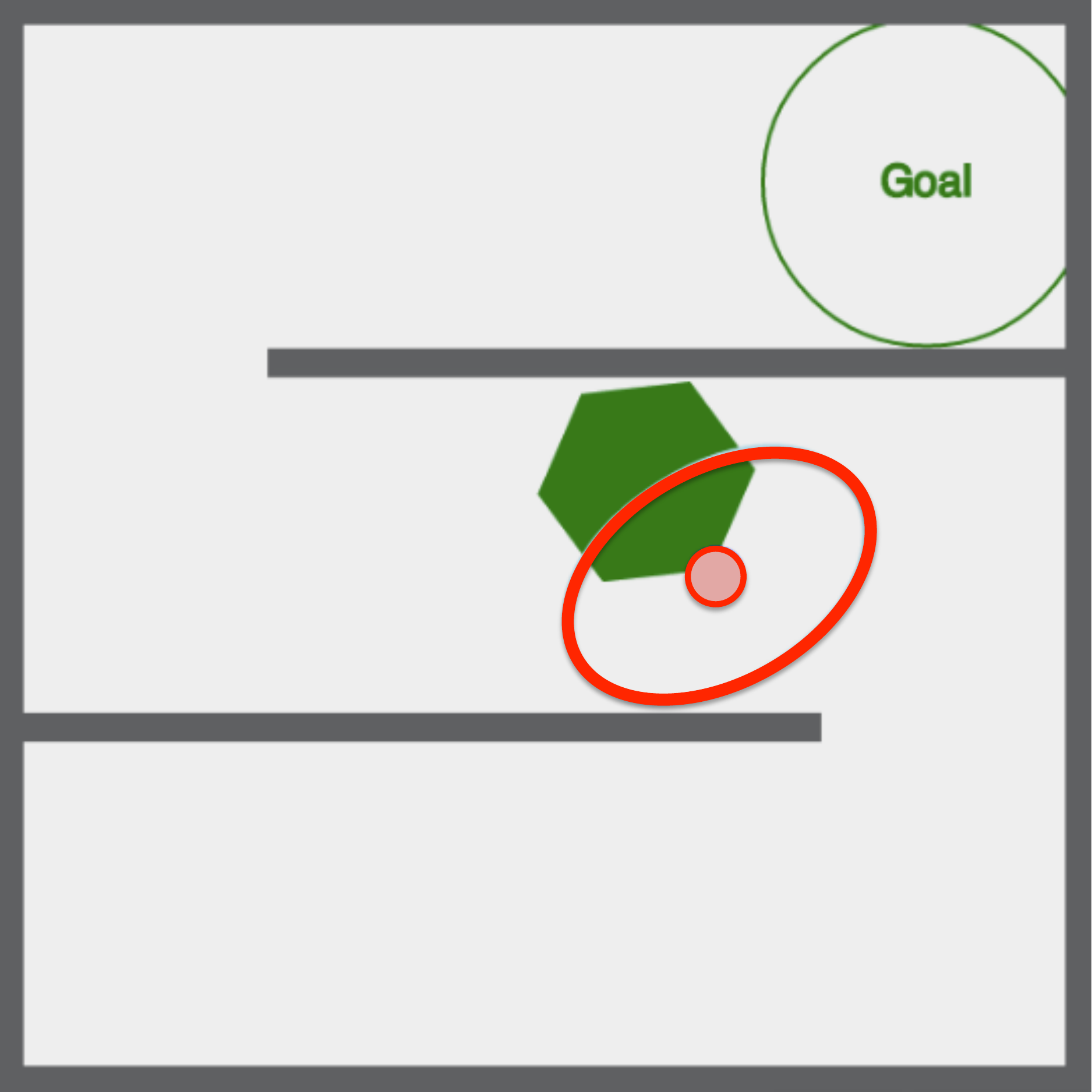}	\put(10,80){\textbf{b} }\end{overpic}~
\begin{overpic}[width =\figwid]{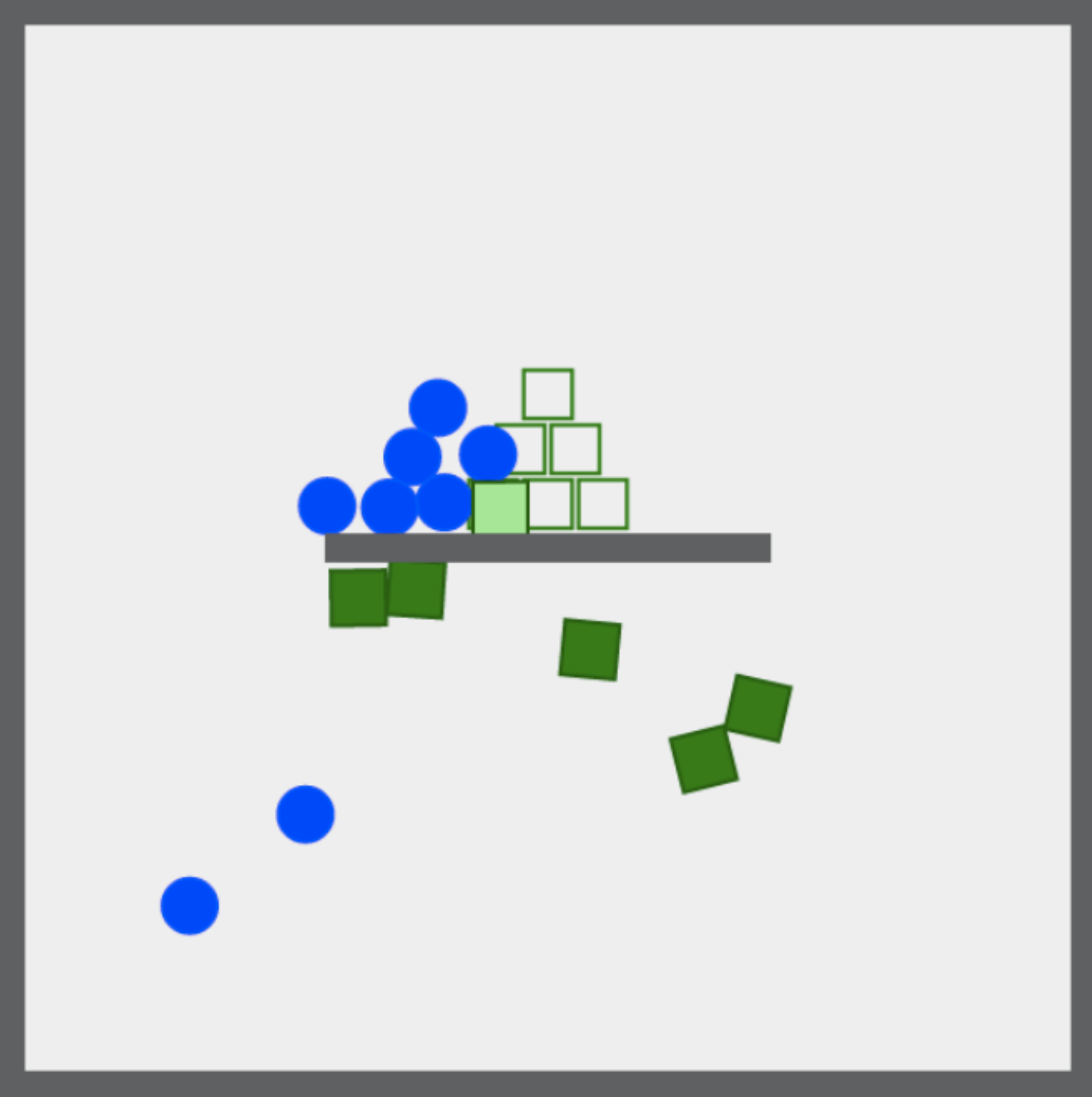}	\put(10,80){\textbf{c} }\end{overpic}\\
%\\
%\begin{overpic}[width =\figwid]{ControlPos.pdf}	\put(10,80){\textbf{d} }\end{overpic}~
%\begin{overpic}[width =\figwid]{VaryControl.pdf}	\put(10,80){\textbf{e} }\end{overpic}~
%\begin{overpic}[width =\figwid]{VaryForage.pdf}	\put(10,80){\textbf{f} }\end{overpic}
\caption{\label{fig:5experiments}
Screenshots from our online experiments controlling multi-particle systems with limited, global control.
\textbf{(a)} Varying the number of particles from 1-500
\textbf{(b)} Comparing 4 levels of visual feedback 
\textbf{(c)} Varying noise from 0 to 200\% of control authority
%\textbf{(d)} Controlling the position of 1 to 10 particles
%\textbf{(e)} Comparing 3 control architectures to assemble
%\textbf{(f)} Comparing 3 control architectures to forage.
}
\end{figure}

%\begin{figure}
%\renewcommand{\figwid}{0.32\columnwidth}
%\subfloat[][Vary Number]{\label{fig:VaryNum}
%\begin{overpic}[width =\figwid]{VaryNum.pdf}\end{overpic}}
%%
%\subfloat[][Vary Visual Feedback]{\label{fig:VaryVis}
%%\begin{overpic}[width =\figwid]{VaryVisFS.pdf}\end{overpic}
%\begin{overpic}[width =\figwid]{VaryVisCH.pdf}\end{overpic}}
%%
%\subfloat[][Vary Noise]{\label{fig:VaryNoise}
%\begin{overpic}[width =\figwid]{VaryNoise.pdf}\end{overpic}}\\
%%
%\subfloat[][Control Position]{\label{fig:ControlPos}
%\begin{overpic}[width =\figwid]{ControlPos.pdf}\end{overpic}}
%%
%\subfloat[][Vary Control: Assembly]{\label{fig:VaryControl}
%\begin{overpic}[width =\figwid]{VaryControl.pdf}\end{overpic}}
%%
%\subfloat[][Vary Control: Foraging]{\label{fig:Forage}
%\begin{overpic}[width =\figwid]{VaryForage.pdf}\end{overpic}}
%%
%\caption{\label{fig:5experiments}
%Screenshots from our online experiments controlling multi-robot systems with limited, global control.
%\textbf{(a)} Varying the number of robots from 1-500
%\textbf{(b)} Comparing 4 levels of visual feedback 
%\textbf{(c)} Varying noise from 0 to 200\% of control authority
%\textbf{(d)} Controlling the position of 1 to 10 robots
%\textbf{(e)} Comparing 3 control architectures to assemble
%\textbf{(f)} Comparing 3 control architectures to forage.
%\vspace{-2em}
%}
%\end{figure}

The goal of these online experiments is to test several scenarios involving large-scale human-swarm interaction (HSI), and to do so with a statistically-significant sample size. Towards this end, we have created \href{http://www.swarmcontrol.net/show_results}{SwarmControl.net}: an open-source, online testing platform suitable for inexpensive deployment and data collection on a scale not yet seen in swarm robotics research. Screenshots from this platform are shown in Fig.~\ref{fig:5experiments}.  \href{https://github.com/crertel/swarmmanipulate.git}{All code} \href{http://www.swarmcontrol.net/show_results}{and experimental results} are online at \cite{Chris-Ertel2016}.

%Our online experiments show that numerous particles responding to global control inputs are directly controllable by a human operator without special training, that the visual feedback of the swarm state should be simple to increase task performance, and that humans perform swarm-object manipulation faster using attractive control schemes than repulsive control schemes.

We developed a flexible testing framework for online human-swarm interaction studies. Over 5,000 humans performed over 20,000 swarm-robotics experiments with this framework, logging almost 700 hours of experiments.
These experiments indicated three lessons used for designing automatic controllers for object manipulation with particle swarms:

1) When the number of particles is large ($>50$) varying the number of particles does not significantly affect the performance.

2) Swarm control is robust to IID noise.

3) Controllers that only use the mean and variance of the swarm can perform better than controllers with full feedback.
%There are two halves to our framework: the server backend and the client-side (in-browser) frontend. The server backend is responsible for tabulating results, serving webpages containing the frontend code, and for issuing unique identifiers to each experiment participant. The in-browser frontend is responsible for running an experiment---that is to say, accepting user input, updating the state of the robot swarm, and ultimately evaluating task completion.

%% wherein we outline the process that a user takes to participate in an experiment
%\subsection{Methods}

\subsection{Implementation}

Our web server generates a unique identifier for each participant and sends it along with the landing page to the participant. 
%This identifier is stored as a browser cookie and is attached to all results the participant generates. 
%The participant's browser prompts for confirmation of the terms-of-service and offers a menu of experiments.
%Once the participant selects an experiment, their browser makes a new request to the server to load the experiment's webpage. The server sends scaffold HTML describing the layout of the page and a script block containing the experiment. 
A script on the participant's browser runs the experiment and posts the experiment data to the server. 
Anonymized human subject data was collected under IRB \#14357-01.

We designed six experiments to investigate human control of large swarms for manipulation tasks.  Screenshots of representative experiment are shown in Fig.~\ref{fig:5experiments}.  Each experiment examined the effects of varying a single parameter: population of particles for manipulation, four levels of visual feedback, different levels of Brownian noise.%, position control with 1 to 10 particles, and three control architectures for both assembly and foraging tasks. 
 The users could choose which experiment to try, and our architecture randomly assigned a parameter value for each trial.  We recorded the completion time and the participant ID for each successful trial.  
%As Fig.~\ref{fig:Learning} shows, one-third of all participants played only a single game.  Still, many played multiple games, and their decreasing completion times demonstrates their skills improved.

\subsection{Varying number}
\begin{figure}
\begin{overpic}[width = 0.9\columnwidth]{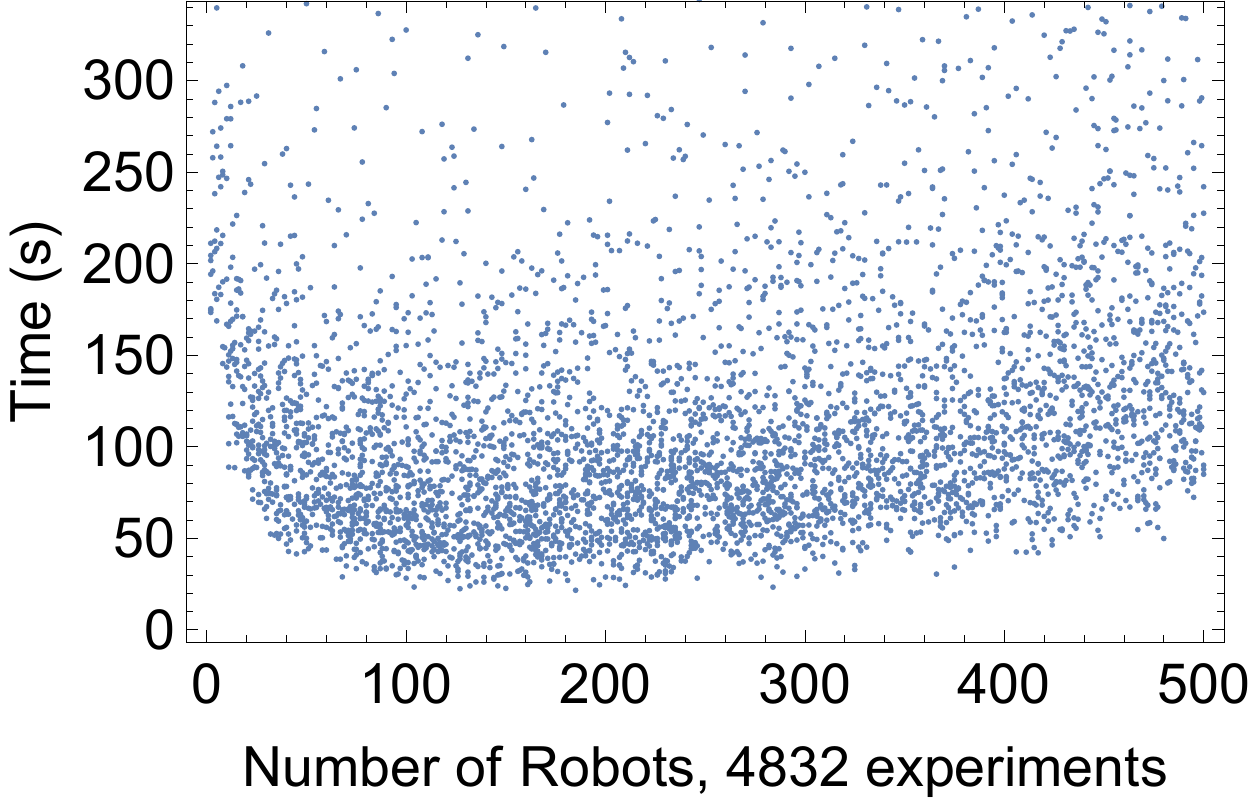}\end{overpic}
%\begin{overpic}[width = 0.48\columnwidth]{measureLearning.pdf}\end{overpic}
\caption{
\label{fig:ResVaryNu}Data from \emph{Varying Number} using particles to push an object through a maze to a goal location. 
% Best-fit linear and quadratic lines are overlaid for comparison. 
%(Right) Skill improves as players retry tests using data from \emph{Varying Number}.
}
\end{figure}

%Transport of goods and materials between points is at the heart of all engineering and construction in real-world systems. 
This experiment varied from 1 to 500 the population of particles used to transport an object. The total area, maximum particle speed, and total net force the swarm could produce were constant. The particles pushed a large hexagonal object through an  {\sffamily S}-shaped maze. We hypothesized participants would complete the task faster with more particles. The results, shown in Fig.~\ref{fig:ResVaryNu}, do not support our hypothesis, indicating a minimum around 130 particles, but only a gradual increase in completion time from 50 to 500.

\subsection{Varying visualization}
\begin{figure}[b!]
\renewcommand{\figwid}{0.24\columnwidth}
\begin{overpic}[width =\figwid]{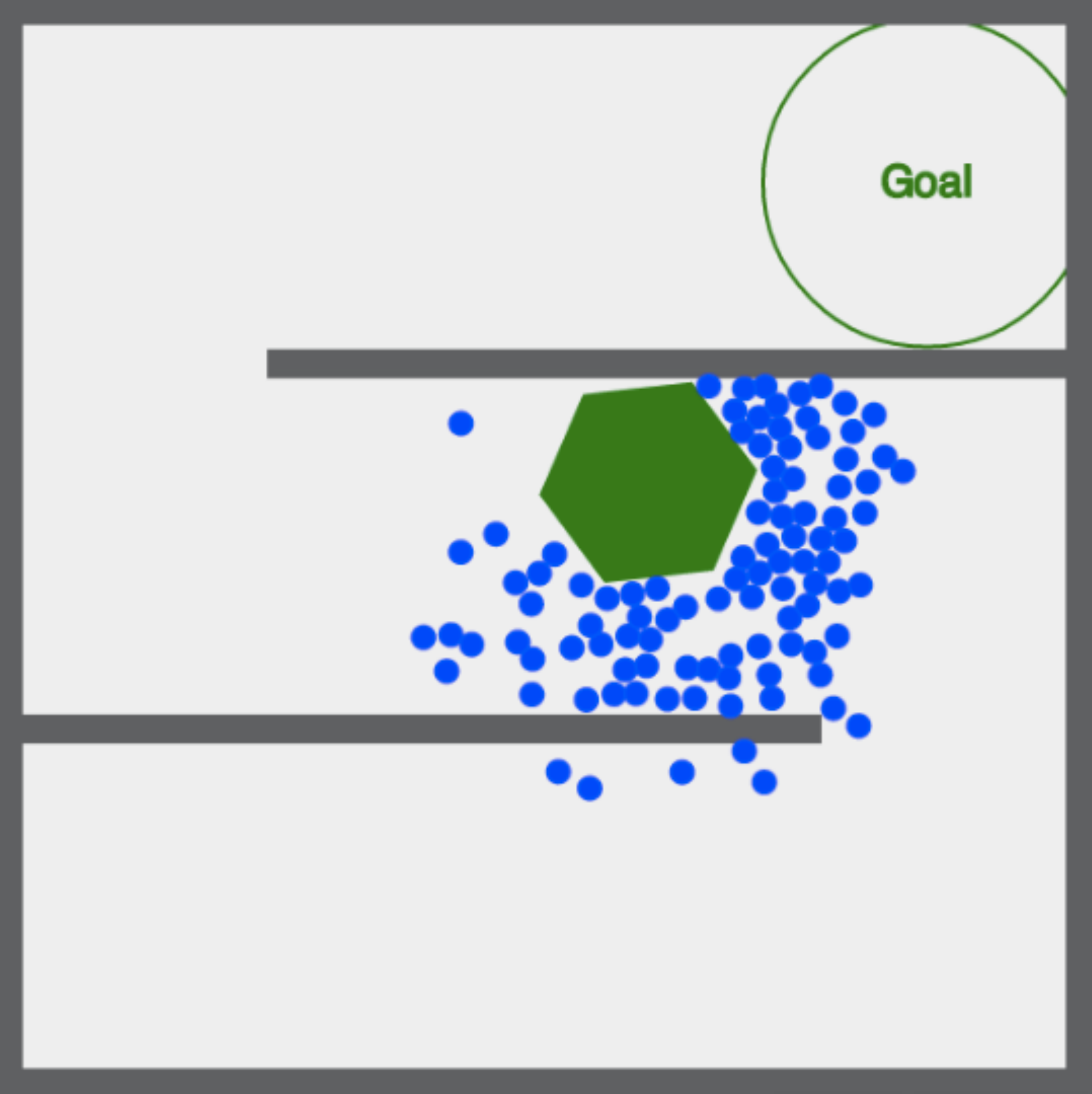}\put(20,15){Full-state}\end{overpic}
\begin{overpic}[width =\figwid]{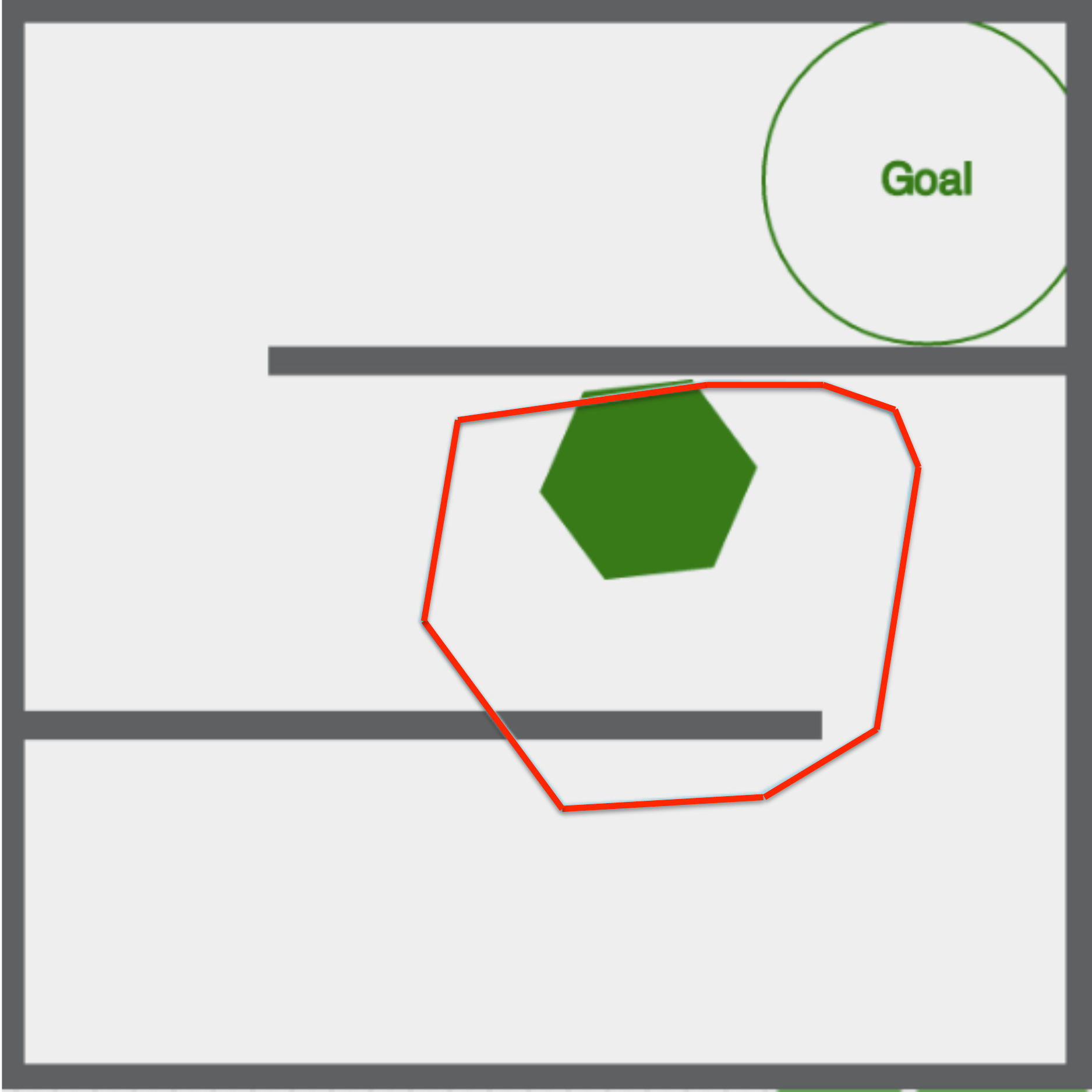}\put(10,15){Convex-hull}\end{overpic}
\begin{overpic}[width =\figwid]{VaryVisMV.pdf}\put(10,15){Mean + var}\end{overpic}
\begin{overpic}[width =\figwid]{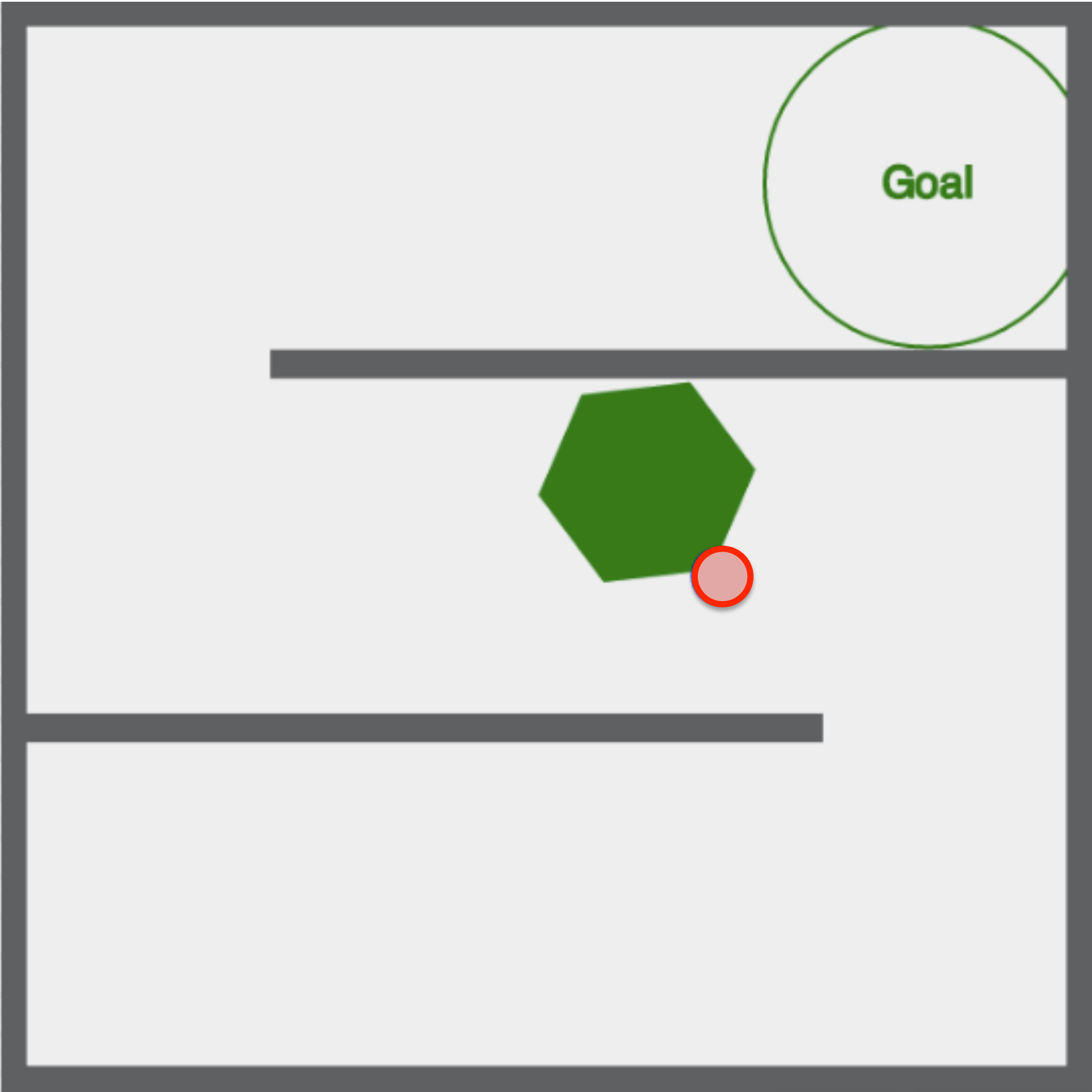}\put(30,15){Mean}\end{overpic}
\vspace{-.5em}
\caption{\label{fig:Visualization}Screenshots from task \emph{Vary Visualization}. This experiment challenges players to quickly steer 100 particles (blue discs) to push an object (green hexagon) into a goal region. We record the completion time and other statistics.
%\vspace{-1em}
}
\end{figure}

\begin{figure}
%  \vspace{-20pt}
  \begin{center}
  \includegraphics[width=0.8\columnwidth]{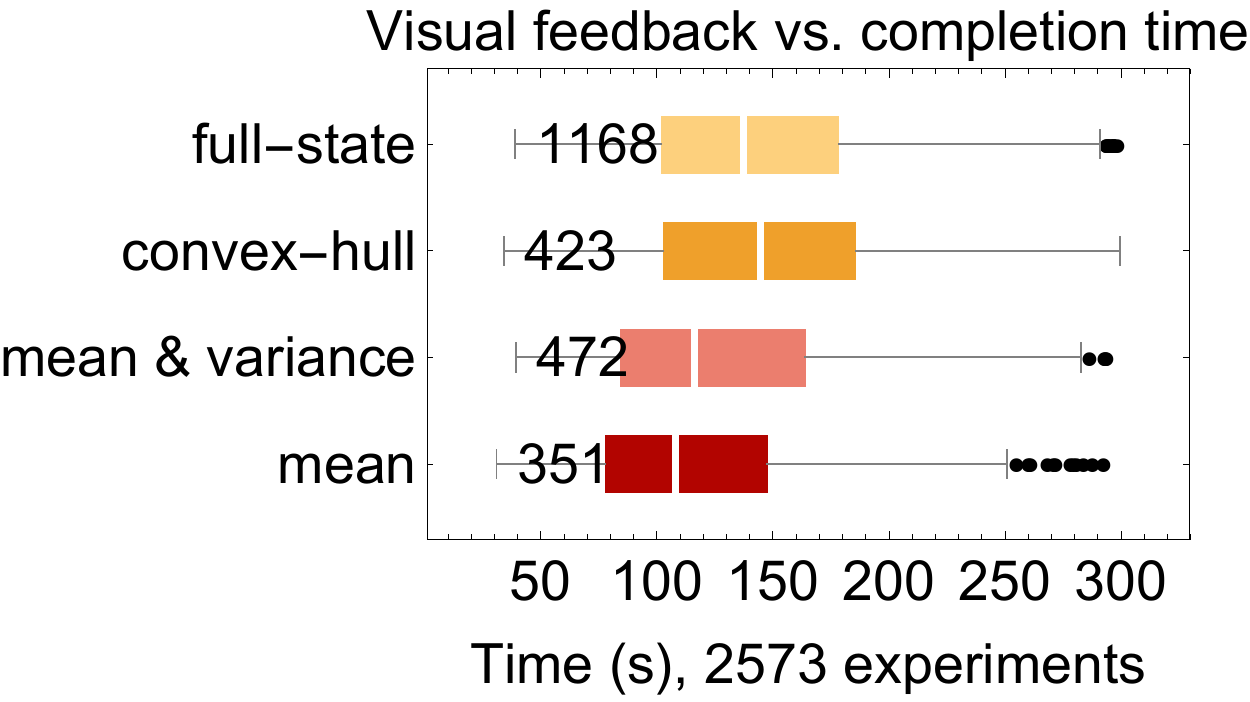}
  \end{center}
%  \vspace{-1em}
\caption{\label{fig:ResVaryVis} Completion-time results for the four levels of visual feedback shown in Fig.~\ref{fig:Visualization}.  Players performed better with limited feedback.
\vspace{-1em}
}
\end{figure}

%Sensing is expensive, especially on the nanoscale. To see nanocars~\cite{Chiang2011} fastened molecules that fluoresce light when activated by a strong light source. 
%Unfortunately, multiple exposures can destroy these molecules, a process called \emph{photobleaching}. 
%Photobleaching can be minimized by lowering the excitation light intensity, but \cite{Cazes2001} showed this increases the probability of missed detections.  
This experiment explores manipulation with varying amounts of sensing information: {\bf full-state} sensing provides the most information by showing the position of all particles; {\bf convex-hull} draws a convex hull around the outermost particles; {\bf mean} provides the average position of the population; and {\bf mean + variance} adds a confidence ellipse. Fig.~\ref{fig:Visualization} shows screenshots of the same particle swarm with each type of visual feedback. Full-state requires $2n$ data points for $n$ particles. Convex-hull requires at worst $2n$, but according to \cite{har2011expected}, the expected number is $O(2 n^{1/3})$.  Mean requires two, and variance three, data points.  Mean and mean + variance are convenient even with millions of particles.

Our hypothesis predicted a steady decay in performance as the amount of visual feedback decreased.
Our experiment indicated the opposite: players with just the mean completed the task faster than those with full-state feedback.  As Fig.~\ref{fig:ResVaryVis}.b shows, the levels of feedback arranged by increasing completion time are [mean, mean + variance, full-state, convex-hull].  All experiments lasting over 300s were removed, under the assumption that the user stopped playing. 
Using ANOVA analysis, we rejected the null hypothesis that all visualization methods are equivalent, with $p$-value $2.69\times10^{-19}$.
Anecdotal evidence from beta-testers who played the game suggests that tracking 100 particles is overwhelming---similar to schooling phenomenons that confuse predators---while working with just the mean + variance is like using a ``spongy'' manipulator. Our beta-testers described convex-hull feedback as confusing and irritating.  A single particle left behind an obstacle will stretch the entire hull, obscuring the majority of the swarm. Similarly, our algorithm must be robust to outliers.
%obscuring what the rest of the swarm is doing.   

\begin{figure}[b!]
\renewcommand{\figwid}{0.49\columnwidth}
\begin{overpic}[width =\figwid]{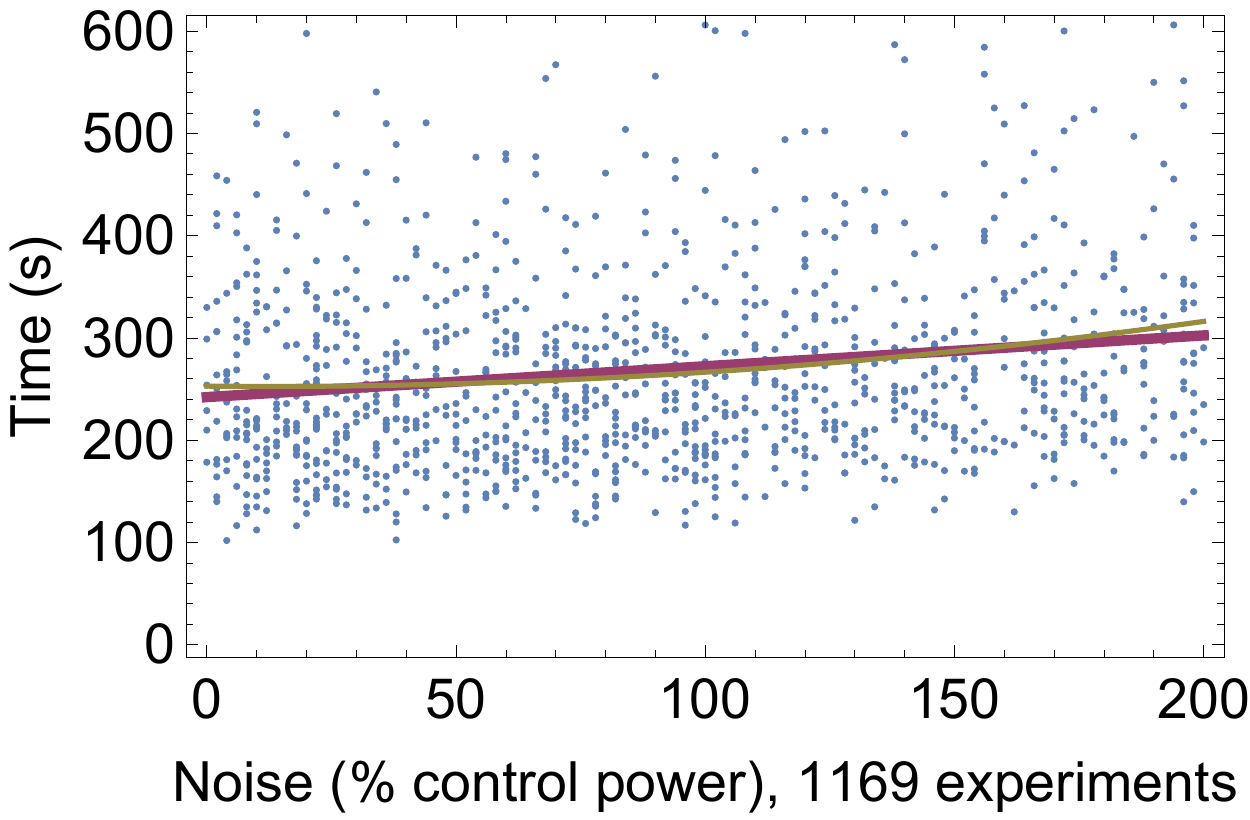}\end{overpic}
\begin{overpic}[width =\figwid]{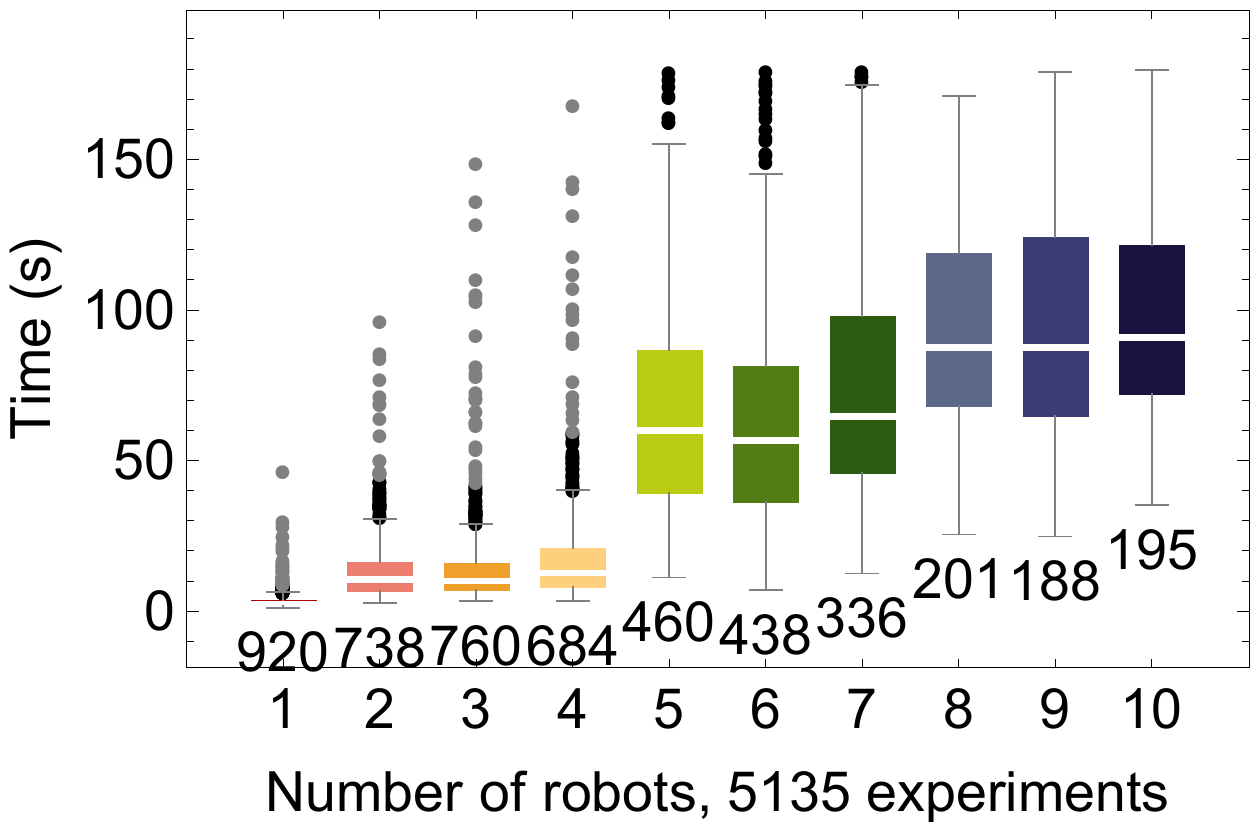}\end{overpic}
%\vspace{-1em}
\caption{\label{fig:ResVaryNoisePosition} Left: Varying the noise from 0 to 200\% of the maximum control input resulted in only a small increase in completion time. Right: For position control, increasing the number of particles resulted in longer completion times.  For more than 4 particles the goal pattern contained a void, which may have caused the jump in completion times.
%\vspace{-1em}
}
\end{figure}

\subsection{Varying noise}
%Micro-particles are affected by random collisions with molecules. The effect of these collisions is called Brownian motion.
This experiment varied the strength of %these 
disturbances to study how noise affects human control of large swarms. Noise was applied at every timestep:
\begin{align*}
\dot{x}_i &= u_x + m_i\cos(\psi_i)\\
 \dot{y}_i &= u_y + m_i\sin(\psi_i).
 \end{align*}
$m_i,\psi_i$ were uniformly IID, with $m_i\in[0,M]$ and $\psi_i\in[0,2\pi]$. $M$ was a constant for each trial ranging from 0 to 200\% of the maximum control power ($u_{\rm{max}}$).
 
We hypothesized 200\% noise was the largest a human could be expected to control---at 200\% noise, the particles move erratically.  Disproving our hypothesis, the results in Fig.~\ref{fig:ResVaryNoisePosition}.a show only a 40\% increase in completion time for the maximum noise. This indicates swarm control is robust to IID noise.

\section{Global Control Laws for a holonomic swarm}
\label{sec:theory}
%%%%%%%%%%%%%%%%%%%%%%%%%%%%%%%%%%%%%%%%%%%%%%%%%%%%%%%%%%%

% Theory:
%% Models (of each robot, of the global input)
%% Impossibility result
%% Controlling Mean proof
%% Controlling Variance proof (CLF)
%% A Hybrid controller using hysteresis

Emboldened by the three lessons from our online experiments, this section presents automatic controllers for large numbers of particles that only rely on the first two moments of the swarm position distribution.

We represent particles as holonomic robots that move in the 2D plane. We want to control position and velocity of the particles. 
First, assume a noiseless system containing one robot with mass $m$.
 Our inputs are global forces $[u_x,u_y]$. We define our state vector $\mathbf{x}(t)$ as the $x$ position, $x$ velocity, $y$ position and $y$ velocity.
The state-space representation in standard form is: 
\begin{align}\label{eq:stdform}
\dot{\mathbf{x}}(t)  &=  A \mathbf{x}(t) + B \mathbf{u}(t).
\end{align}

and our state space representation is:
\begin{equation}
\begin{bmatrix}
\dot{x}\\ 
\ddot{x}\\
\dot{y}\\
\ddot{y}
\end{bmatrix} = \begin{bmatrix}
0 & 1 & 0 & 0 \\
0 & 0 & 0 & 0\\
0 & 0 & 0 & 1\\
0 & 0 & 0 & 0
\end{bmatrix}  \begin{bmatrix}
x\\
\dot{x}\\
y\\
\dot{y}
\end{bmatrix} + \begin{bmatrix}
0 & 0 \\
\frac{1}{m} & 0 \\
0 & 0 \\
0 & \frac{1}{m}
\end{bmatrix} 
 \begin{bmatrix}
 u_x\\
 u_y\end{bmatrix}.
\end{equation}

We want to find the number of states that we can control, which is given by the rank of the \emph{controllability matrix}
\begin{align}
\mathcal{C} &= [ B, AB, A^2B, ... , A^{n-1}B ].\\
\textrm{Here }
\mathcal{C}&=\left[
\begin{matrix} 
0 & 0\\
\frac{1}{m} & 0 \\
0 & 0 \\
0 & \frac{1}{m}
\end{matrix}
\,\middle\vert\,
\begin{matrix} 
\frac{1}{m}& 0\\
0 & 0\\
0 & \frac{1}{m}\\
0 & 0
\end{matrix}
\,\middle\vert\,
\begin{matrix} 
0 & 0\\
0 & 0 \\
0 & 0 \\
0 & 0
\end{matrix}
\,\middle\vert\,
\begin{matrix} 
0 & 0\\
0 & 0\\
0 & 0\\
0 & 0
\end{matrix}
 \right],\\
 \quad \textrm{rank}(\mathcal{C})&=4,
\end{align}
and thus all four states are controllable. This section starts by proving independent position control of many robots is not possible, but the mean position can be controlled. We then provide conditions under which the variance of many robots is also controllable.

\subsection{Independent control of many particles is impossible}
In this model, a single particle is fully controllable. For holonomic robots, movement in the $x$ and $y$ coordinates are independent, so for notational convenience without loss of generality we will focus only on movement in the $x$ axis. Given $n$ particles to be  controlled in the $x$ axis, there are $2n$ states: $n$ positions and $n$ velocities. Without loss of generality, assume $m=1$.
Our state-space representation is:
\begin{equation}
\begin{bmatrix}
\dot{x}_{1}\\ 
\ddot{x}_{1}\\
\vdots\\
\dot{x}_{n}\\
\ddot{x}_{n}

\end{bmatrix} = \begin{bmatrix}
0 & 1 & \ldots & 0 & 0 \\
0 & 0 & \ldots& 0 & 0 \\
\vdots &  \vdots & \ddots & \vdots & \vdots \\
0 & 0  & \ldots & 0 & 1 \\
0 & 0 & \ldots& 0 & 0 
\end{bmatrix}  \begin{bmatrix}
x_{1}\\
\dot{x}_{1}\\
\vdots \\
x_{n}\\
\dot{x}_{n}
\end{bmatrix} + \begin{bmatrix}
0\\
1\\
\vdots\\
0\\
1
\end{bmatrix} u_x.
\end{equation}
 However, just as with one particle, we can only control two states because the controllability matrix $\mathcal{C}_n$ has rank two:
\begin{equation}
\mathcal{C}_n=\left[ \begin{matrix} 
0\\
1\\
\vdots\\
0\\
1
\end{matrix}
\,\middle\vert\,
  \begin{matrix} 
1\\
0\\
\vdots\\
1\\
0
\end{matrix}
\,\middle\vert\,
\begin{matrix} 
0\\
0\\
\vdots\\
0\\
0
\end{matrix}
\,\middle\vert\,
\begin{matrix} 
0\\
0\\
\vdots\\
0\\
0
\end{matrix}\,\middle\vert\,
\ldots \right], \quad \textrm{rank}(\mathcal{C}_n)=2.
\end{equation}  
\subsection{Controlling the mean position}\label{sec:controlMeanPosition}
This means \emph{any} number of particles controlled by a global command have just two controllable states in each axis. We cannot arbitrarily control the position and velocity of two or more robots, but have options on which states to control.  %One option is to control the position and velocity of the $j^{th}$ robot. To find a potentially more useful option, 
We create the following reduced order system that represents the mean $x$ position and velocity of the $n$ particles:
\begin{align}
\begin{bmatrix}\nonumber
\dot{\bar{x}} \\
\ddot{\bar{x}}
\end{bmatrix} &= \frac{1}{n} \begin{bmatrix}
0& 1& \ldots &0& 1 \\
0& 0& \ldots &0& 0
\end{bmatrix}
\begin{bmatrix}
x_{1}\\
\dot{x}_{1}\\
\vdots\\
x_{n}\\
\dot{x}_{n}
\end{bmatrix} \\
&+ \frac{1}{n}\begin{bmatrix}
0& 0&  \ldots &0& 0 \\
0& 1&  \ldots &0& 1
\end{bmatrix}\begin{bmatrix} 
0\\
1\\
\vdots\\
0\\
1
\end{bmatrix} u_x.
\end{align}
Thus:
\begin{equation}
\begin{bmatrix}
\dot{\bar{x}} \\
\ddot{\bar{x}}
\end{bmatrix} = \begin{bmatrix}
0& 1 \\
0& 0
\end{bmatrix}
\begin{bmatrix}
\bar{x}\\
\dot{\bar{x}}
\end{bmatrix} + \begin{bmatrix} 
0\\
1
\end{bmatrix} u_x.
\end{equation}

We again analyze the controllability matrix $\mathcal{C}_{\mu}$:
\begin{equation}
\mathcal{C}_\mu=\left[ \begin{matrix} 
0\\
1
\end{matrix}
\,\middle\vert\,
 \begin{matrix} 
1\\
0
\end{matrix}
 \right],  \quad \textrm{rank}(\mathcal{C}_\mu)=2.
\end{equation}
Thus the mean position and mean velocity are controllable.

%Due to symmetry of the control input, only the mean position and mean velocity are controllable. However, 
There are several techniques for breaking the symmetry of the control control input to allow controlling more states, for example by using obstacles as in \cite{Becker2013b}, or by allowing independent noise sources as in \cite{beckerIJRR2014}.

We control mean position with a PD controller that uses the mean position and mean velocity. $[u_x,u_y]^\top$ is the global force applied to each robot:
\begin{align}
u_x &= K_{p}(x_{\rm{goal}} - \bar{x}) + K_{d}(0-\dot{\bar{x}}), \nonumber\\
u_y &= K_{p}(y_{\rm{goal}}  - \bar{y}) + K_{d}(0-\dot{\bar{y}}).  \label{eq:PDcontrolPosition}
\end{align}
 $K_{p}$ is the proportional gain, and $K_{d}$ is the derivative gain.

\subsection{Controlling the variance}\label{sec:VarianceControl}

The variance, $\sigma_x^2,\sigma_y^2$, of $n$ robots' position is computed as:
\begin{align}\label{eq:meanVar}
 \overline{x}(\mathbf{x}) = \frac{1}{n} \sum_{i=1}^n x_{i}, \qquad  %\nonumber \\ 
\sigma_x^2(\mathbf{x}) &= \frac{1}{n} \sum_{i=1}^n (x_{i} - \overline{x})^2,  \nonumber \\ 
 \overline{y}(\mathbf{x}) = \frac{1}{n} \sum_{i=1}^n y_{i}, \qquad  %\nonumber \\ 
\sigma_y^2(\mathbf{x}) &= \frac{1}{n} \sum_{i=1}^n (y_{i} - \overline{y})^2.  
%NOT 1/(N-1) because we are measuring the actual Variance, not the variance of samples drawn from a distribution
\end{align}

Controlling the variance requires being able to increase and decrease the variance.  We will list a sufficient condition for each. 
Microscale systems are affected by unmodelled dynamics. 
These unmodelled dynamics are dominated by Brownian noise, described by~\cite{einstein1956investigations}. 
To model this~\eqref{eq:stdform} must be modified as follows:
\begin{align}
\dot{\mathbf{x}}(t)  &=  A \mathbf{x}(t) + B \mathbf{u}(t) + W \bm{\varepsilon}(t),
\end{align}
where $W\bm{\varepsilon}(t)$ is a random perturbation produced by Brownian noise with magnitude $W$. Given a large obstacle-free workspace with $\mathbf{u}(t)= 0$, a \emph{Brownian noise} process increases the variance linearly with time.
\begin{equation}
\dot{\sigma}_x^2(\mathbf{x}(t), \mathbf{u}(t))  = W \bm{\varepsilon},
\quad \sigma_x^2(t)  = \sigma_x^2(0) + W \bm{\varepsilon} t.
\end{equation}
 If faster dispersion is needed, the swarm can be pushed through obstacles such as a diffraction grating or Pachinko board as in~\cite{Becker2013b}. 

If robots with radius $r$ are in a bounded environment with sides of length $[\ell_x, \ell_y]$, the unforced variance asymptotically grows to the variance of a uniform distribution,
\begin{align}
[\sigma_x^2,\sigma_y^2] = \frac{1}{12}[ (\ell_x - 2 r)^2,(\ell_y - 2 r)^2].\label{eq:VarianceUniformDistribution}
\end{align}

 A flat obstacle can be used to decrease variance. Pushing a group of dispersed robots against a flat obstacle will decrease their variance until the minimum-variance (maximum density) packing  is reached. For large $n$, ~\cite{graham1990penny} showed that the minimum-variance packing  for $n$ circles with radius $r$ is 
 \begin{align} \label{eq:optimalvar}
 \sigma^2_{\rm{optimal}}(n,r) \approx   \frac{\sqrt{3}}{\pi} n r^2\approx 0.55 n r^2.
 \end{align} 
% Sloan minimized second moment U = 1/d^2 \sum_{i=1}^{n} || P_i - P ||^2 = (\sqrt{3} n^2)/(4 \pi), where d = 2r.
% Therefore the variance is d^2 (\sqrt{3} n^2)/(4 \pi), or 4^r^2 (\sqrt{3} n^2)/(4 \pi) = r^2 (\sqrt{3} n^2)/(\pi)If we only need to reduce variance in a single axis, the variance can be reduced to zero, given sufficient space.

We will prove the goal is globally asymptotically stabilizable by using a control-Lyapunov function, as in \cite{Lyapunov1992}.  A suitable Lyapunov function is the squared variance error:
\begin{align}
\label{eq:LyapunovVariance}
V(t,\bf{x})  &= \frac{1}{2} (\sigma^2(\mathbf{x}) - \sigma^2_{\rm{goal}})^2,\nonumber\\
\dot{V}(t,\bf{x}) &= (\sigma^2(\mathbf{x})-\sigma^2_{\rm{goal}})\dot{\sigma}^2(\mathbf{x}).
\end{align}
We note here that $V(t,\mathbf{x})$ is positive definite and radially unbounded, and $V(t,\mathbf{x}) \equiv 0$ only at $\sigma^2(\mathbf{x}) = \sigma^2_{\rm{goal}}$.
To make $\dot{V}(t,\mathbf{x})$ negative semi-definite, we choose
\begin{align}\label{eq:controlVariance}
u(t) &=   \begin{cases}
	 \mbox{move to wall} &\mbox{if } \sigma^2(\mathbf{x})>\sigma^2_{\rm{goal}} \\ 
	 \mbox{move from wall} & \mbox{if } \sigma^2(\mathbf{x}) \le \sigma^2_{\rm{goal}}.
\end{cases} 
\end{align}
 For such a $u(t)$,
 \begin{align}
\dot{\sigma}^2(\mathbf{x}) &=   \begin{cases}
	 \mbox{negative} &\mbox{if } \sigma^2(\mathbf{x})> \max(\sigma^2_{\rm{goal}}, \sigma^2_{\rm{optimal}}(n,r))  \\ 
	 W \epsilon & \mbox{if } \sigma^2(\mathbf{x}) \le \sigma^2_{\rm{goal}},
\end{cases} 
\end{align} and thus
$\dot{V}(t,{\bf x})$ is negative definite and the variance is globally asymptotically stabilizable.% \hfill$\blacksquare$ 

A PID controller to regulate the variance to $\sigma^2_{\rm{ref}}$ is:
\begin{align}
u_x &= K_{p}(x_{\rm{goal}}(\sigma^2_{\rm{ref}}) - \bar{x}) - K_{d}\bar{v}_x + K_{i}(\sigma^2_{\rm{ref}}-\sigma^2_{x}), \nonumber\\
u_y &= K_{p}(y_{\rm{goal}}(\sigma^2_{\rm{ref}})  - \bar{y}) - K_{d}\bar{v}_y + K_{i}(\sigma^2_{\rm{ref}}-\sigma^2_{y}).  \label{eq:PDcontrolVariance}
\end{align}
We call the gain scaling the variance error $K_i$ because the variance, if unregulated, integrates over time.
Eq.~\eqref{eq:PDcontrolVariance} assumes the nearest wall is to the left of the robot at $x=0$, and chooses a reference goal position that in steady-state would have the correct variance according to \eqref{eq:VarianceUniformDistribution}:
\begin{align}
x_{\rm{goal}}(\sigma^2_{\rm{ref}}) = \ell_x/2 = r + \sqrt{3\sigma^2_{\rm{ref}}}.
\end{align}
 If a wall to the right is closer, the signs of $[K_p,K_i]$ are inverted, and the location $x_{\rm{goal}}$ is translated.

\subsection{Controlling both mean and variance}

The mean and variance of the swarm cannot be controlled simultaneously, however if the dispersion due to Brownian motion is less than the maximum controlled speed, we can adopt the hybrid, hysteresis-based controller shown in Alg.~\ref{alg:MeanVarianceControl} to regulate the mean and variance.  Such a controller normally controls the mean position, but switches to minimizing variance if the variance exceeds some $\sigma_{\rm{max}}^2$. 
 Variance is reduced until less than $\sigma_{\rm{min}}^2$, then control again regulates the mean position. 
 This technique satisfies control objectives that evolve at different rates as in~\cite{kloetzer2007temporal}, and the hysteresis avoids rapid switching between control modes. The process is illustrated in Fig.~\ref{fig:hysteresis}.

\begin{algorithm}
\caption{Hybrid mean and variance control}\label{alg:MeanVarianceControl}
\begin{algorithmic}[1]
\Require Knowledge of swarm mean $[\bar{x},\bar{y}]$, variance $[\sigma_x^2, \sigma_y^2]$, the locations of the rectangular boundary $\{x_{\rm{min}}, x_{\rm{max}}, y_{\rm{min}}, y_{\rm{max}}\}$, and the target mean position $[x_{\rm{target}},y_{\rm{target}}]$.%TODO: use  \AND, \OR, \XOR, \NOT, \TO, \TRUE, \FALSE \gets
\State $x_{goal} \gets  x_{\rm{target}}$, $y_{\rm{goal}} \gets y_{\rm{target}}$
\Loop
%\State  Compute $\sigma_x^2, \sigma_y^2$

\If {$\sigma_x^2 > \sigma_{\rm{max}}^2$}
\State $x_{\rm{goal}}  \gets x_{\rm{min}}$
\ElsIf { $\sigma_x^2 < \sigma_{\rm{min}}^2$}
\State $x_{\rm{goal}}  \gets  x_{\rm{target}}$
\EndIf

\If{$\sigma_y^2 > \sigma_{\rm{max}}^2$}
\State $y_{\rm{goal}}  \gets y_{\rm{min}}$
\ElsIf { $\sigma_y^2 < \sigma_{\rm{min}}^2$}
\State $y_{\rm{goal}}  \gets  y_{\rm{target}}$
\EndIf
\State Apply \eqref{eq:PDcontrolPosition} to move toward $[x_{\rm{goal}}, y_{\rm{goal}}]$
\EndLoop
\end{algorithmic}
\end{algorithm}

\begin{figure}
\centering
\begin{overpic}[width = 0.8\columnwidth]{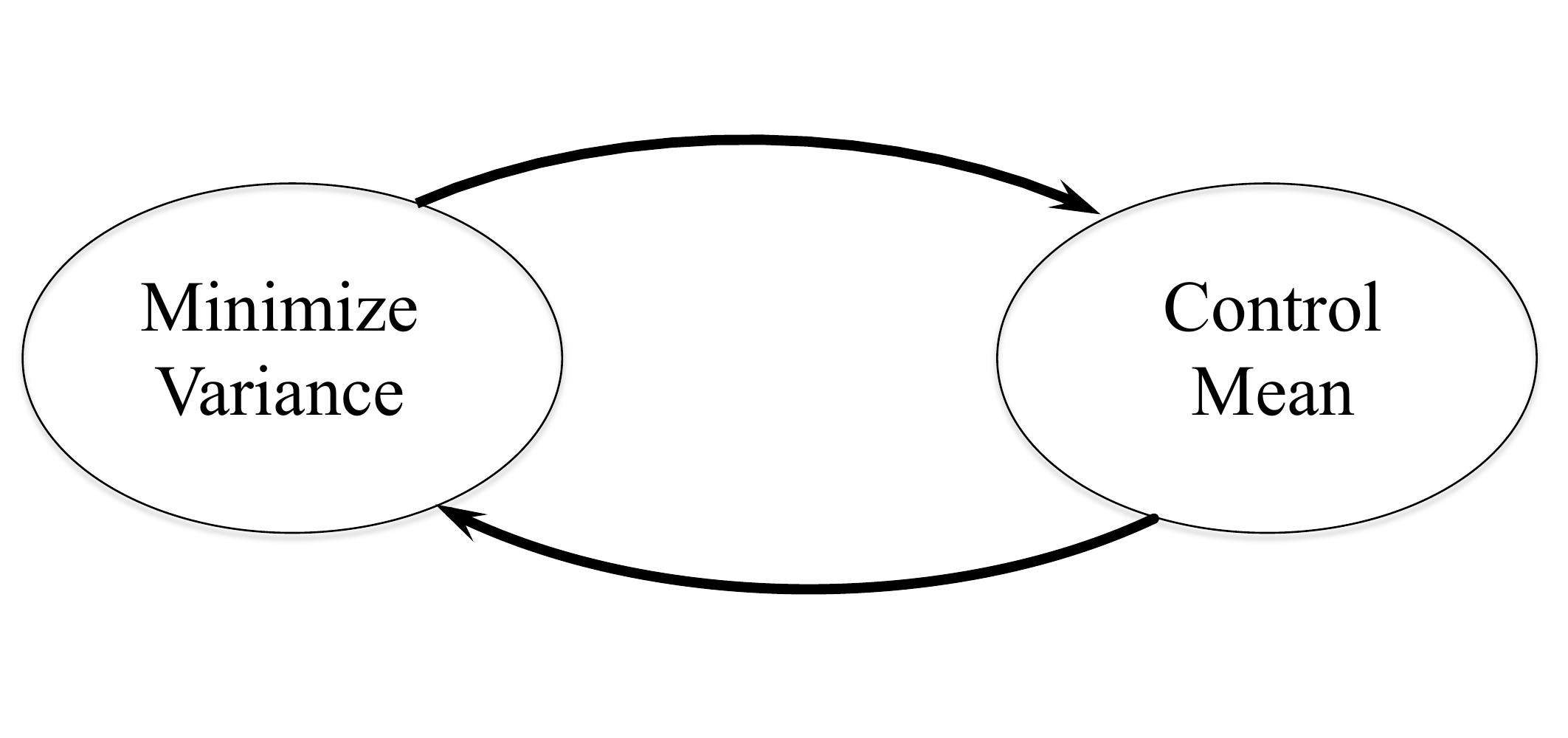}
\put(33,40){$\sigma^2(\mathbf{x}) < \sigma^2_{\rm{min}}$ }
\put(33,15){$\sigma^2(\mathbf{x}) > \sigma^2_{\rm{max}}$}\end{overpic}
%\begin{overpic}[width = 0.45\columnwidth]{VarianceMinMaxBand2v2.pdf}\end{overpic}
\vspace{-0.5em}
\caption{\label{fig:hysteresis}  Hysteresis to control swarm mean and variance. 
%(Right) Switching conditions for variance control are set as a function of $n$, and designed to be larger than the optimal packing density. The plot uses robot radius $r=1/10$.
%\vspace{-2em}
}
\end{figure}

%\begin{figure}
%\centering
%\begin{overpic}[width = 0.5\columnwidth]{VarianceMinMaxBand2v2.pdf}\end{overpic}
%\vspace{-1em}
%\caption{\label{fig:VarianceMinMaxBand} The switching conditions for variance control are set as a function of $n$, and designed to be larger than the optimal packing density. The above plot uses robot radius $r=1/10$.
%}\vspace{-1em}
%\end{figure}

A key challenge is to select proper values for $\sigma_{\rm{min}}^2$ and $\sigma_{\rm{max}}^2$.  The optimal packing variance was given in \eqref{eq:optimalvar}.
The random packings generated by pushing our robots into corners are suboptimal, so we choose the conservative values: 
%shown in Fig.~\ref{fig:hysteresis}:
\begin{align} \label{eq:sigMaxMin}
 \sigma^2_{\rm{min}} &= 2.5r+ \sigma^2_{\rm{optimal}}(n,r), \nonumber\\
  \sigma^2_{\rm{max}} &= 15r+ \sigma^2_{\rm{optimal}}(n,r).
  \end{align}
%\sigma \approx 0.371258 n r,   \sigma_{min} = n*r*1/2, \sigma_{min} = n*r*3/2, 
%  this result was verified in http://www-math.mit.edu/~tchow/penny.pdf

%$\sigma^2_{optimal}(n,r)$ for $n$ circles with radius $r$ is $\approx 
 %    \frac{\sqrt{3}}{\pi} (n r)^2\approx 0.55(n r)^2 $. 
% Sloan minimized second moment U = 1/d^2 \sum_{i=1}^{n} || P_i - P ||^2 = (\sqrt{3} n^2)/(4 \pi), where d = 2r.
% Therefore the variance is d^2 (\sqrt{3} n^2)/(4 \pi), or 4^r^2 (\sqrt{3} n^2)/(4 \pi) = r^2 (\sqrt{3} n^2)/(\pi)If we only need to reduce variance in a single axis, the variance can be reduced to zero, given sufficient space.

%what images should I show here?
%hysteresis control law, \cite{sadra2014}

\section{Simulation of control laws}\label{sec:simulation}
% Simulation:
%%  Mean control results   (image and plot(s) )
%%  Variance Control      (image and plot(s) )
%%  Hybrid control   (image and plot(s) )

Our simulations use a Javascript port of \href{http://box2d.org/}{Box2D}, a popular 2D physics engine with support for rigid-body dynamics and fixed-time step simulation, presented in \cite{catto2010box2d}.  All experiments in this section ran on a Chrome web browser on a 2.6 GHz Macbook.  \href{https://github.com/aabecker/SwarmControlSandbox/blob/master/exampleControllers/BlockPushingIROS2015.html}{All code is available at}~\cite{Shahrokhi2016blocksimulations}.

\subsection{Controlling the mean position}
\begin{figure}
\centering
\begin{overpic}[width = \columnwidth ]{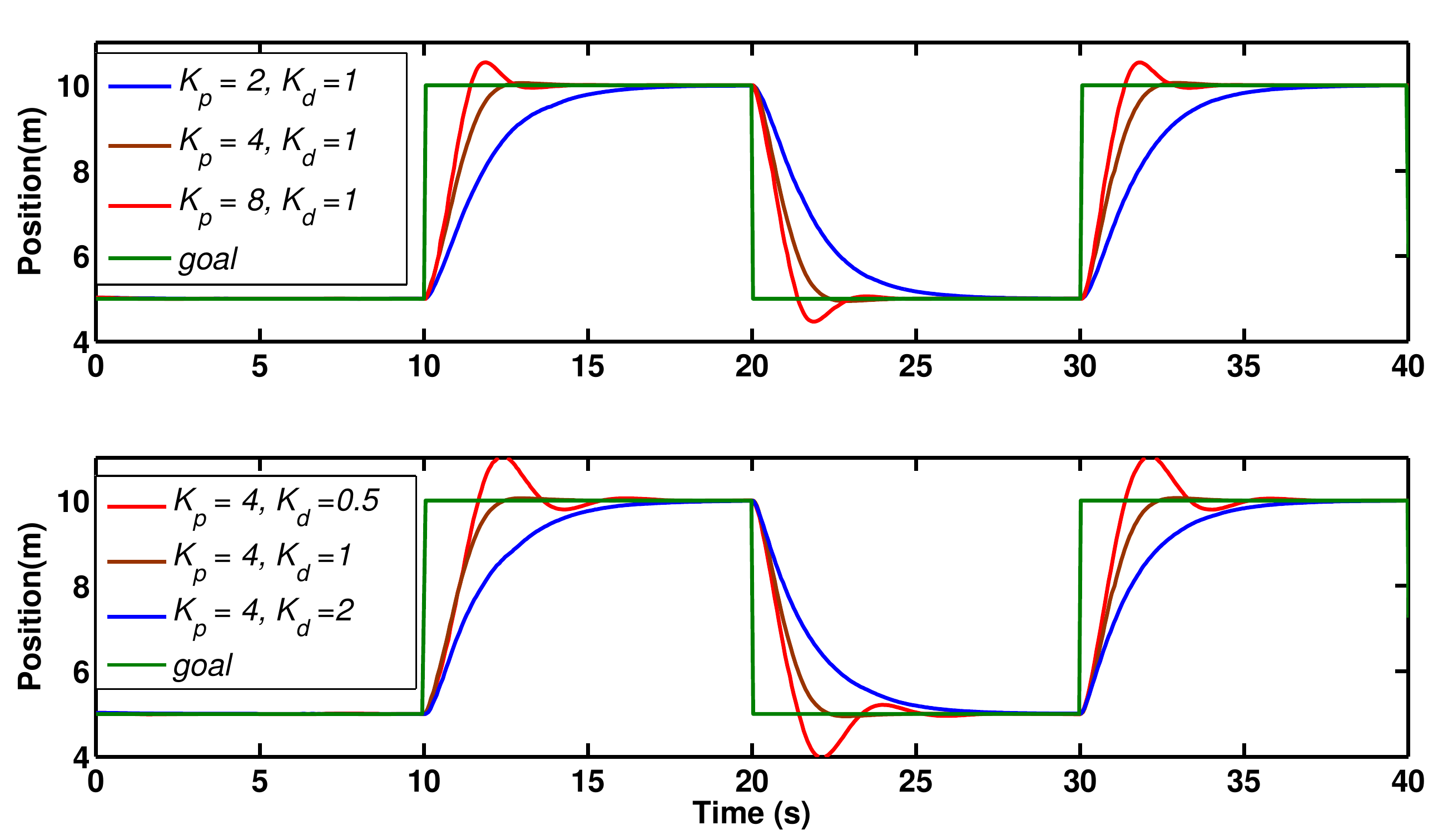}
\end{overpic}
%\begin{overpic}[width =0.49\columnwidth]{meanVariance4.eps} 
%\end{overpic} 
\vspace{-1em} 
\caption{\label{fig:gainvalues} In simulation, tuning proportional ($K_p$, top) and derivative ($K_d$, bottom)  gain values in \eqref{eq:PDcontrolPosition} improves performance with $n = 100$ particles. %b) Simulation result with 100 robots under hybrid control Alg.~\ref{alg:MeanVarianceControl}, which  controls both the mean position (top) and variance (bottom). For ease of analysis, only $x$ position and variance are shown.
%\vspace{-2em}
}
\end{figure}
We performed a parameter sweep using the PD controller \eqref{eq:PDcontrolPosition} to identify the best control gains.  Representative experiments are shown in Fig.~\ref{fig:gainvalues}. 100 particles were used and the maximum speed was 3 meters per second. As shown in Fig.~\ref{fig:gainvalues}, we can achieve an overshoot of 1\% and a  rise time of 1.52~s with $K_{p}= 4$, and  $K_{d} = 1$.

%\begin{figure}
%\centering
%\begin{overpic}[width = \columnwidth* 2/3 ]{meanVideoSnapShot.png}
%\end{overpic}
%\vspace{0 em}
%\caption{\label{fig:meanVideo} A frame from video attachment showing mean position control on a swarm of 200 robots. The mean position of the swarm traces ``SWARM"~\cite{ShivaVideo2015}. 
%%\vspace{-2em}
%}
%\end{figure}

\subsection{Controlling the variance}
\begin{figure}
\centering
\begin{overpic}[width = \columnwidth] {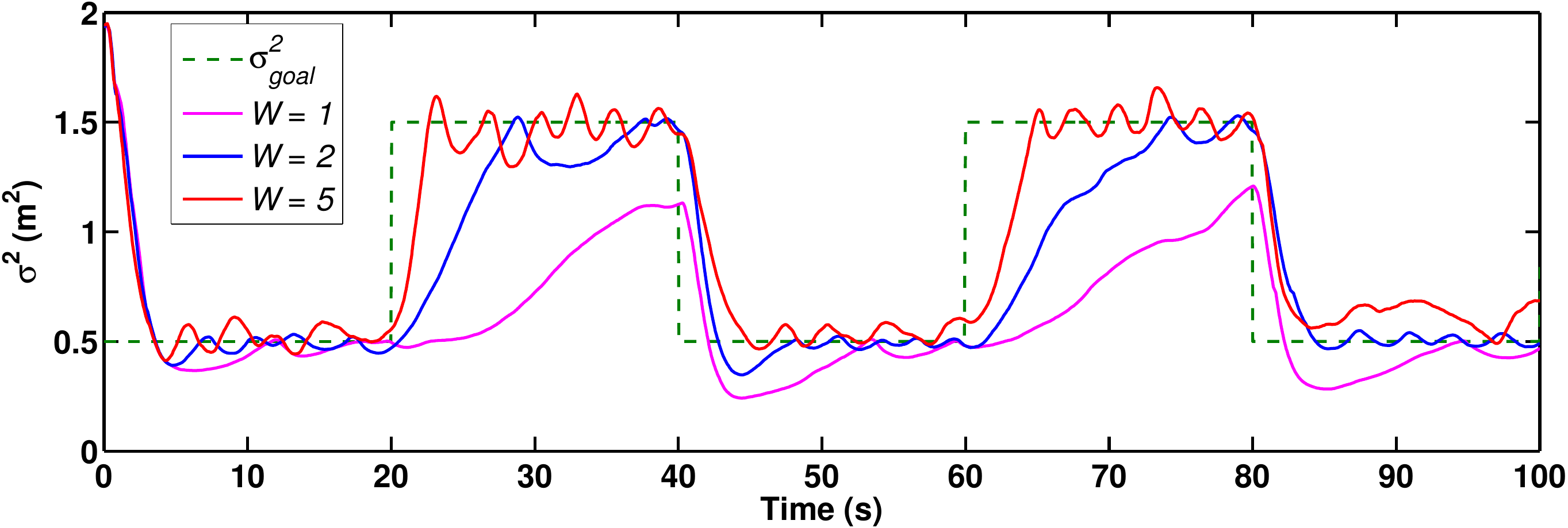}
\end{overpic}
\vspace{-1em}
\caption{\label{fig:varyBrownian} In simulation, increased noise results in more responsive variance control because stronger Brownian noise causes a faster increase of variance.
%\vspace{-2em}
}
\end{figure}

%cite the control law, explain experiment (number of robots, maximum speed, ).

For variance control we use the control law \eqref{eq:PDcontrolVariance}. 
 Results are shown in Fig.~\ref{fig:varyBrownian}, with $K_{p,i,d} = [4,1,1]$.

%\todo{image showing control x variance and y-variance out of phase}

\subsection{Hybrid control of mean and variance}

Fig.~\ref{fig:hybrid} shows a simulation run of the hybrid controller in Alg.~\ref{alg:MeanVarianceControl} with 100 particles in a square workspace containing no internal obstacles. 
%\todo{plot showing 1.5 cycles of mean position, and a variance goal.  We might need a longer time}
\begin{figure}
\centering
\begin{overpic}[width = \columnwidth]{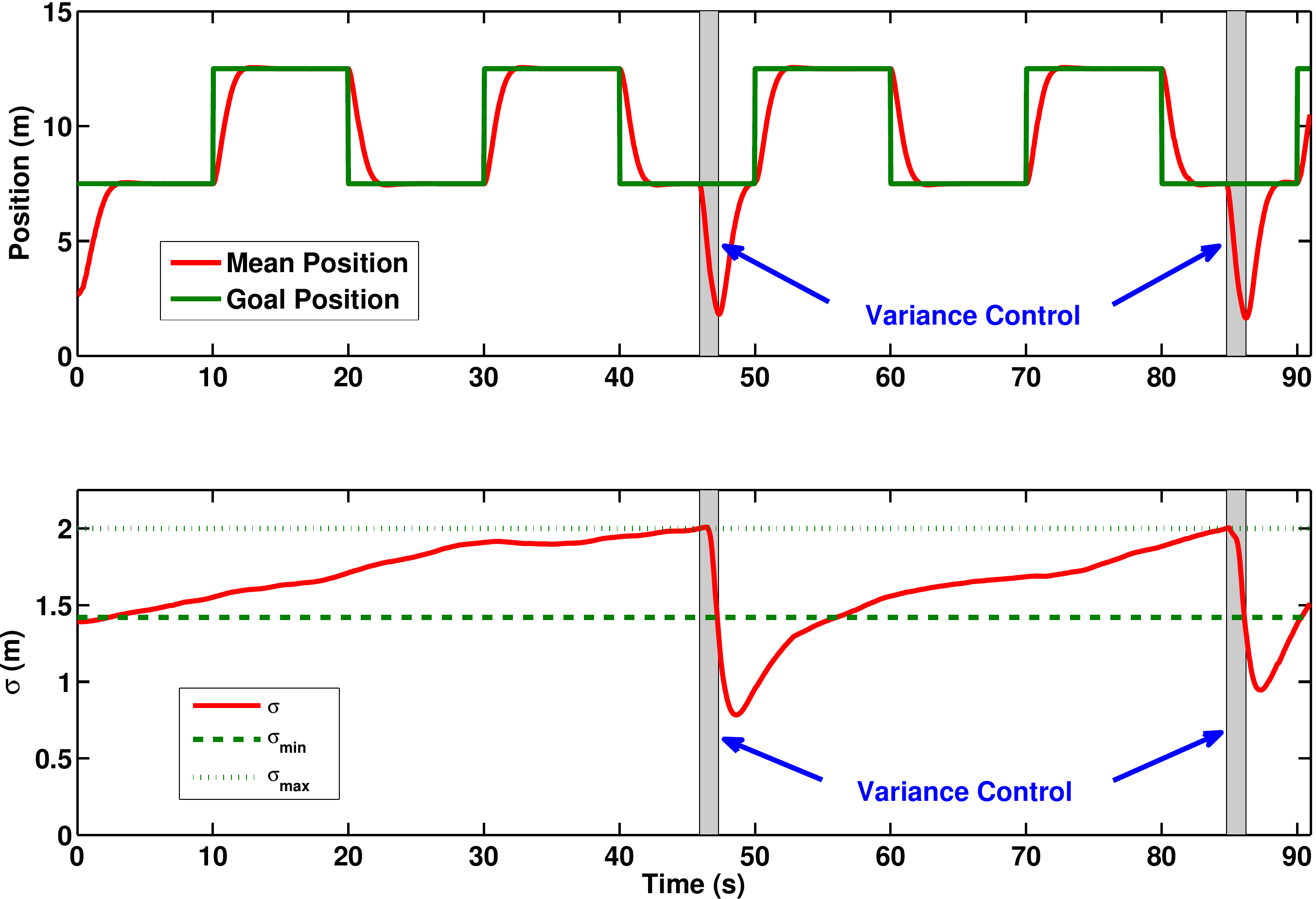}
\end{overpic}
\vspace{-1em}
\caption{\label{fig:hybrid} Simulation result with 100 particles under hybrid control Alg.~\ref{alg:MeanVarianceControl}, which  controls both the mean position (top) and variance (bottom). For ease of analysis, only $x$ position and variance are shown.
\vspace{-2em}
}
\end{figure}

%\begin{figure}
%\centering
%\begin{overpic}[width = \columnwidth * 2/3]{VideoSnapShot.png}
%\end{overpic}
%\vspace{-1em}
%\caption{\label{fig:videoVar} A frame from video, using Alg.~\ref{alg:MeanVarianceControl} to control variance and mean position of a swarm of 200 robots~\cite{ShivaVideo2015}.
%%\vspace{-2em}
%}
%\end{figure}

%%%%%%%%%%%%%%%%%%%%%%%%%%%%%%%%%%%%%%%%%%%%%%%%%%%%%%%%%%%
\section{Particle Swarm Object manipulation}\label{sec:exp}
%%%%%%%%%%%%%%%%%%%%%%%%%%%%%%%%%%%%%%%%%%%%%%%%%%%%%%%%%%%

This section analyzes an \emph{object manipulation} task attempted by our hybrid, hysteresis-based controllers. The swarm must deliver the object to the goal region.  To solve this object manipulation task we divide the task into three components: 1) designing a policy for the object, 2) pushing the object with a compliant swarm, and 3) managing outliers.

The table below summarizes the simulation results for each 10 successful trials:
\begin{center}
\begin{tabular}{ c c }
\hline
Method & Result, mean$\pm$std (s)\\ [0.5ex] 
\hline \hline
Value Iteration (VI) & 367 $\pm$ 253 \\ 
\hline
VI + Potential Fields (PF) & 271 $\pm$ 267 \\  
\hline
VI + Outlier Rejection (OR) & 245 $\pm$ 135\\
\hline
BFS + PF + OR & 183 $\pm$ 179 \\
\hline
VI + PF + OR & 90 $\pm$ 35 \\[0.5ex]
\hline
\end{tabular}
\end{center}
\subsection{Learning a policy for the object}\label{subsec:objectpolicy}

To design the policy we first discretize the environment. 
In \cite{ShahrokhiIROS2015}, we used breadth-first search (BFS) on this discretized grid, but using workspace BFS fails to account for the hull of the object and will suggest moves that can cause collisions with the workspace. A configuration-space BFS approach avoids that problem but still fails to model uncertain actuation of the object by the swarm.

To solve both these problems, this paper models object movement as a Markov Decision Process (MDP) with non-deterministic movement.  
  Value iteration,  as described in \cite{Thrun2005}, is used to learn an \emph{optimal policy}.
   At each state the object can be commanded to move in one of eight directions with a small probability of moving in a wrong direction. 
 
% A corresponding reward function gives a high reward to the goal state,  
% a large negative reward to states including obstacles, and a small negative reward to all other states.
The reward function $r(x,\mathbf{u})$ is defined as
\begin{align}
r(x,\mathbf{u}) &=  \left\{
\begin{array}{ll}
     +100, &  \textrm{if } \mathbf{u} \textrm{ leads to goal state}\\
      -100, & \textrm{if } \mathbf{u} \textrm{ leads to an obstacle state} \\
      -1, & \textrm{otherwise}\\
\end{array} 
\right.
\end{align}
 where $x$ is the current state and $\mathbf{u}$ is the action.   %cumulative expected
  Value iteration computes $\hat{V}(x)$, the expected discounted sum reward if the optimal policy is implemented, for the object starting in each state $x$. The optimal policy is %the control action that, in probability, results in the largest value:
   \begin{align} \mathbf{D}(x) = \arg\max_{\mathbf{u}}   [ r(x,\mathbf{u}) + \sum\limits_{j=1}^N \hat{V}(x_j) p(x_j|x, \mathbf{u})].  \label{eq:OptimalPolicy}
   \end{align}
   The value function $\hat{V}(x_j) $ is calculated by computing the value $\hat{V}$ for all $N$ states and iterating until convergence:
\begin{align}
\text{for }&\text{$i=1$ to $N$ do} \nonumber \\
&\hat{V} (x_j) = \gamma \max_{\mathbf{u}} [r(x_j,\mathbf{u}) + \sum\limits_{i=1}^N \hat{V}(x_i) p(x_i| x_j,\mathbf{u})] \label{eq:ValueIteration} \nonumber\\
\text{end}& 
\end{align}
In our experiments $\gamma = 0.97$, and \eqref{eq:ValueIteration} was iterated 200 times. A {\sc Matlab} implementation of this algorithm is available at \cite{Becker2015MDP}.
%Rather than a shortest distance to goal, $\nabla \mathbf{M}_{MDP}$  gives the approximate minimum cost to goal.

$\mathbf{M}_{\rm{BFS}}$ and the value function are shown in Fig.~\ref{fig:BFSGradient}. 
In 10 simulations with 100 particles, pushing the object to goal using BFS required 183$\pm$179 s while Policy Iteration required 90$\pm$35 s (mean$\pm$std).
%\todo{(report results of 10 simulations with BFS and with PI.  Give mean and std)}

\begin{figure}
\centering
\renewcommand{\figwid}{3cm}
\begin{overpic}[height=\figwid]{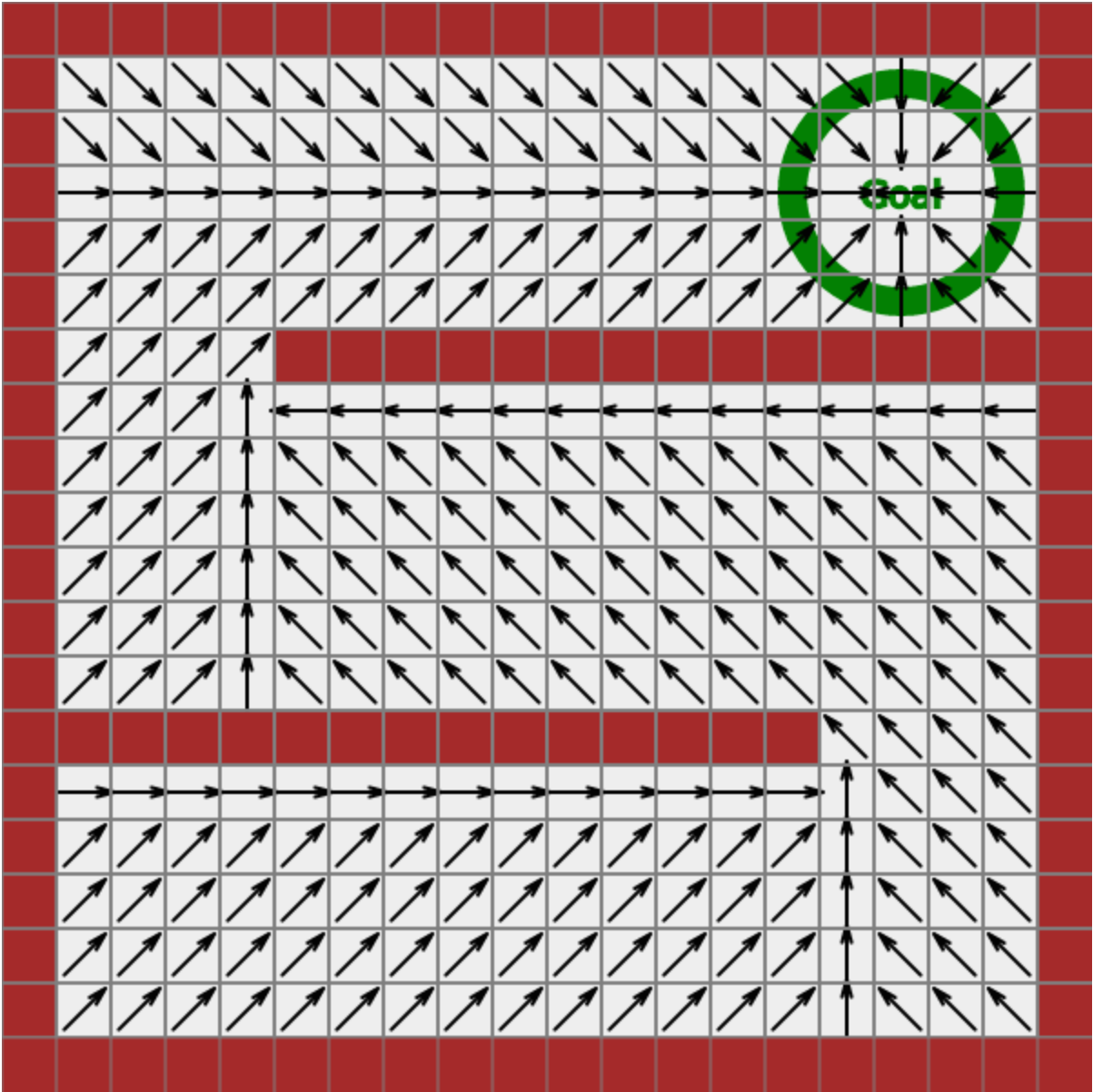}
\end{overpic}
\begin{overpic}[height=\figwid]{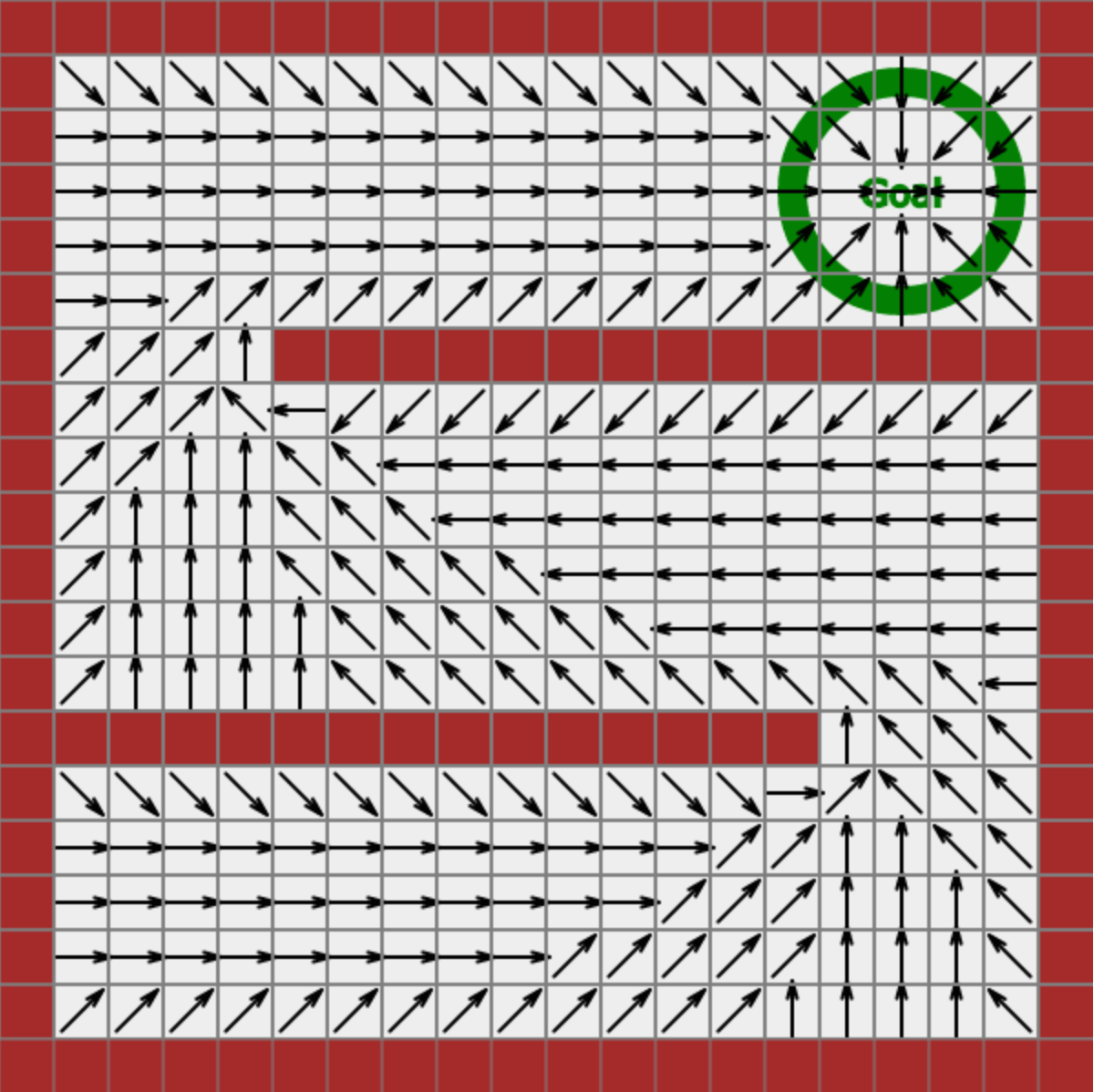}
\end{overpic}
\begin{overpic}[height=\figwid]{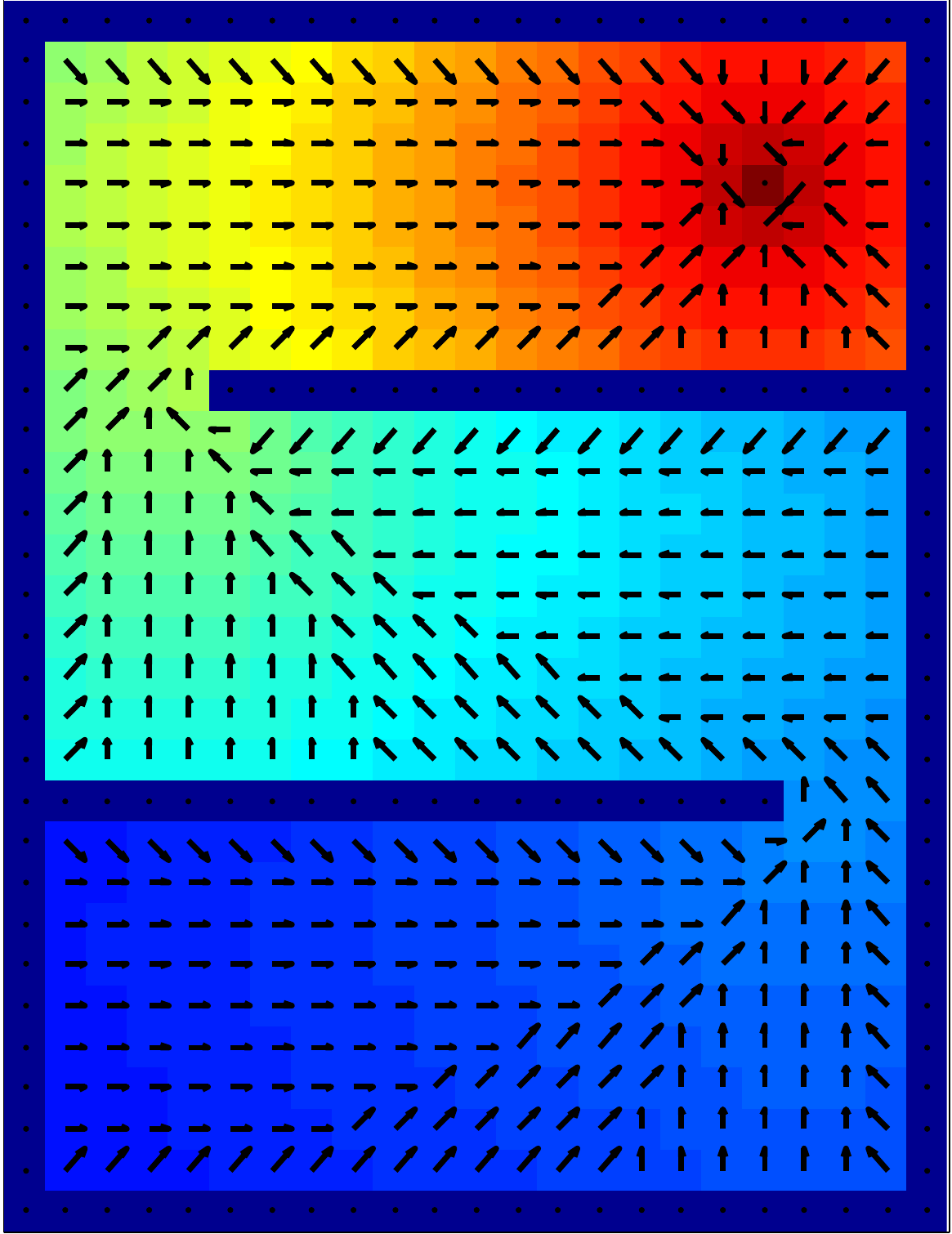}
\end{overpic}
\vspace{-0.5em}
\caption{\label{fig:BFSGradient} BFS finds the shortest path for the moveable object to compute gradient vectors (left). Modeling as an MDP enables encoding penalties for being near obstacles. (Middle) The control policy from value iteration. (Right) The vision algorithm detects obstacles in the hardware setup. This map is used to produce the value function and control policy shown. 
%\vspace{-2em}
}
\end{figure}

\subsection{Potential fields for swarm management with a compliant manipulator}

When the swarm is in front of the object, control law \eqref{eq:PDcontrolPosition} pushes the object backwards.  To fix this, we implement a potential field approach inspired by \cite{spong2008robot} that attracts the swarm to the intermediate goal, but repulses the swarm from in front of the object.
The repulsive potential field is centered at the object's COM and is active for a radius $\rho_0$, but is implemented only when the swarm mean is within $\theta$ of the desired direction of motion $\mathbf{D}(\mathbf{b})$ as shown in Fig.~\ref{fig:potentialField}b.
\begin{align}
F_{\rm{att}} &= -\zeta \Delta \rho / \rho, \\
%\mathbf{u} &= [\bar{x},\bar{y}] - \mathbf{b} \nonumber \\
\phi &=\cos^{-1}\left( \frac{ \mathbf{D}(\mathbf{b}) \cdot ( [\bar{x},\bar{y}] - \mathbf{b}) }{\norm{\mathbf{D}(\mathbf{b})} \norm{ [\bar{x},\bar{y}] - \mathbf{b}} } \right), \\
%\theta &= \mathrm{angularDistance}\left( \alpha, \beta \right)\nonumber \\
 F_{\rm{rep}} &=  \left\{
\begin{array}{ll}
      \eta( 1/\rho- 1/\rho_0) \frac{1}{\rho^2} \Delta \rho, ~& \rho\leq \rho_0~\wedge~\phi <  \theta \\
      0, & \textrm{ otherwise} \\
\end{array},
\right.\\
F_{\rm{pot}} &= F_{\rm{att}} + F_{\rm{rep}} \label{eq:potentialfield}.
\end{align}

In simulations, $\theta =  \pi/2$,  $\eta  = 75$, $\zeta = 2$ and $\rho_0 = 3$. Because the kilobots have a slower time constant, they use $\theta =  \pi/2$,  $\eta  = 50$, $\zeta = 1$ and $\rho_0 = 7.5$. 

In 10 simulations with 100 particles, pushing the object to goal without a repulsive potential field failed in two of twelve runs. No failures occurred with the repulsive potential field.  Of successful trials, completion time without repulsive potential fields required 245$\pm$135 s while using repulsive potential fields required 90$\pm$35 s (mean$\pm$std).
%In 10 simulations with 100 robots, completion time without repulsive potential fields required $\mu=245, \sigma=135$s while using repulsive potential fields required $\mu=90, \sigma=35$.

%\todo{report results of 10 simulations with and without Potential fields.  Give mean and std}

\begin{figure}
\centering
\begin{overpic}[width=1\columnwidth]{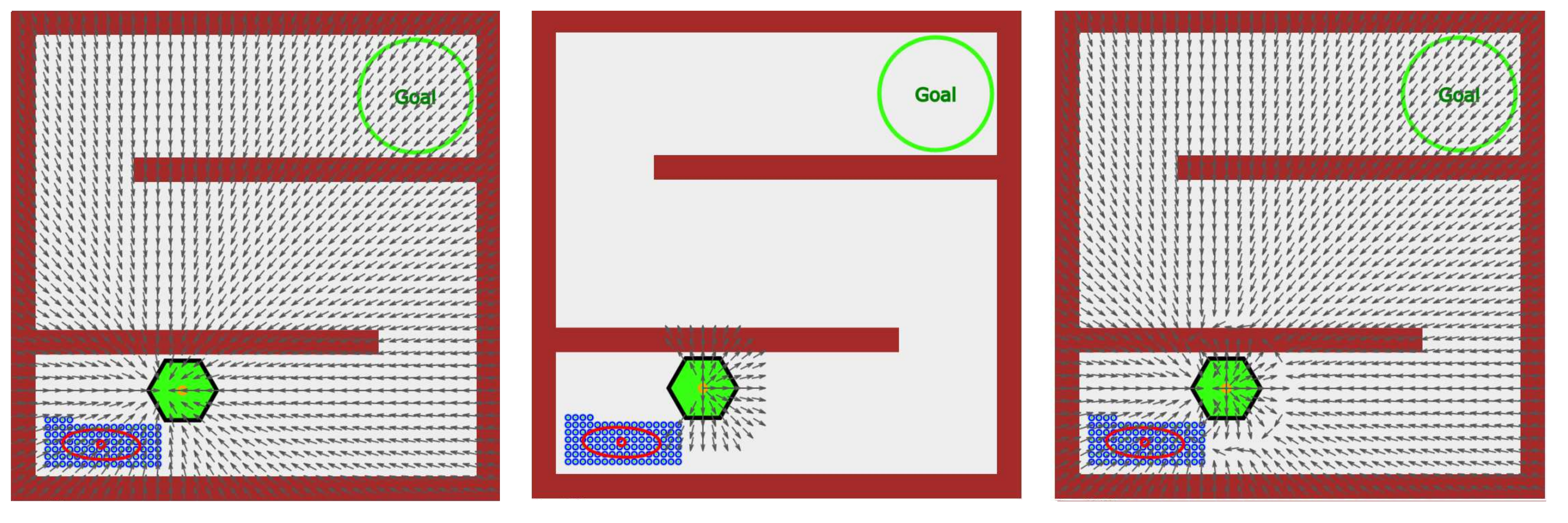}\end{overpic}
%\todo{I like the 'target' symbol, but it is not self-documenting.  We need a legend explaining the min and max variance ellipses, the goal region, the variance, the mean, the object COM, and the target mean position.  I think these are easiest to make in powerpoint.
%Please use the same color and line style for the variance min and max as you use in Figure 4.
%}
%{blockpushingImageWithMeanAndVarianceOverlay.png}
\caption{\label{fig:potentialField} (Left) The attractive field is centered behind the object's COM. (Middle) The repulsive field is centered at the object's COM. (Right) Combining these forces prevents the swarm from pushing the object backwards.}
\end{figure}

\subsection{Outlier rejection}\label{subsec:OutlierRejection}

The variance controller in Alg.~\ref{alg:MeanVarianceControl} is a greedy algorithm that is susceptible to outliers. 
The controller in \cite{ShahrokhiIROS2015} failed in $14\%$ trials, some particles were unable to reach the object because workspace obstacles were blocking them. This failure rate increases if  object weight increases or ground-robot friction increases. The mean and covariance calculations \eqref{eq:meanVar} included all particles in the workspace. Particles that cannot reach the object due to obstacles skew these calculations. The state machine in Fig. \ref{fig:Region}a solves this problem by creating two states for the maze: either main or transfer. Each state has a set of regions representing a discretized visibility polygon. Whenever the object crosses a region boundary the state toggles. The \emph{main} regions are generated by extending obstacles until they meet another obstacle. The \emph{transfer} regions are perpendicular to obstacle boundaries, and act as a buffer between two main regions.

Fig.\ \ref{fig:Region}b shows the regions for the main state. The object is in region 1. An indicator function is applied to \eqref{eq:meanVar} so only robots inside region 1 are counted.  This filtering increases experimental success because the mean calculation only includes nearby robots that can directly interact with the object. 
When the object leaves main region 1 the state switches to transfer. The transfer regions are shown in Fig.\ \ref{fig:Region}c.  The object is in transfer region 1, so only robots in transfer region 1 are included in the mean and covariance calculations. 
 The robots should push the object to the left. Without filtering using regions, the red circle is the mean and the algorithm would instruct the robots to push the object up. The black circle shows the filtered mean and the algorithm instructs the robots to push the object directly left.

In 10 simulations with 100 robots, completion time without outlier rejection required 271$\pm$267 s while using outlier rejection required 90$\pm$35 s (mean$\pm$std).
%\todo{report results of 10 simulations with and without regions.  Give mean and std}

\begin{figure}
\begin{center}
	\begin{overpic}[width=0.9\columnwidth]{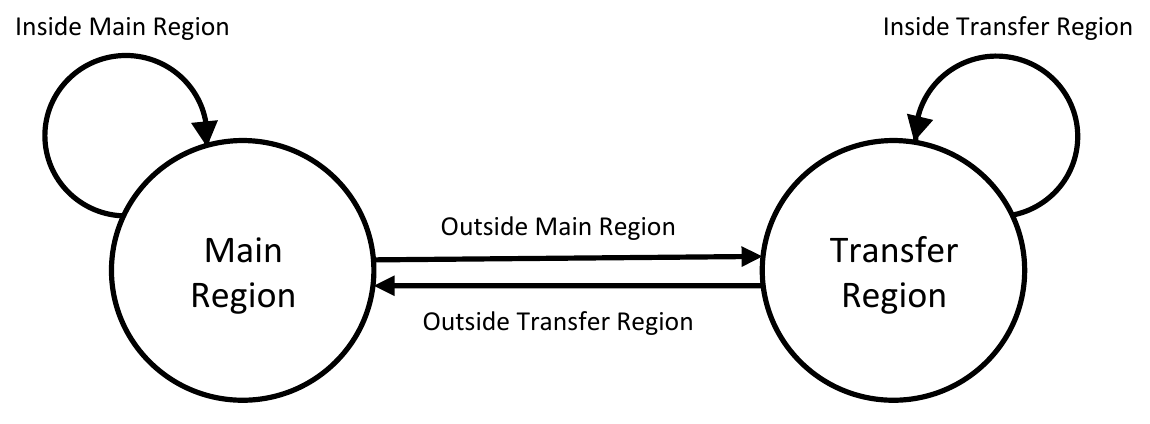}\put(-4,25){\textbf{a}}\end{overpic}\\
	~~\begin{overpic}[width=0.45\columnwidth]{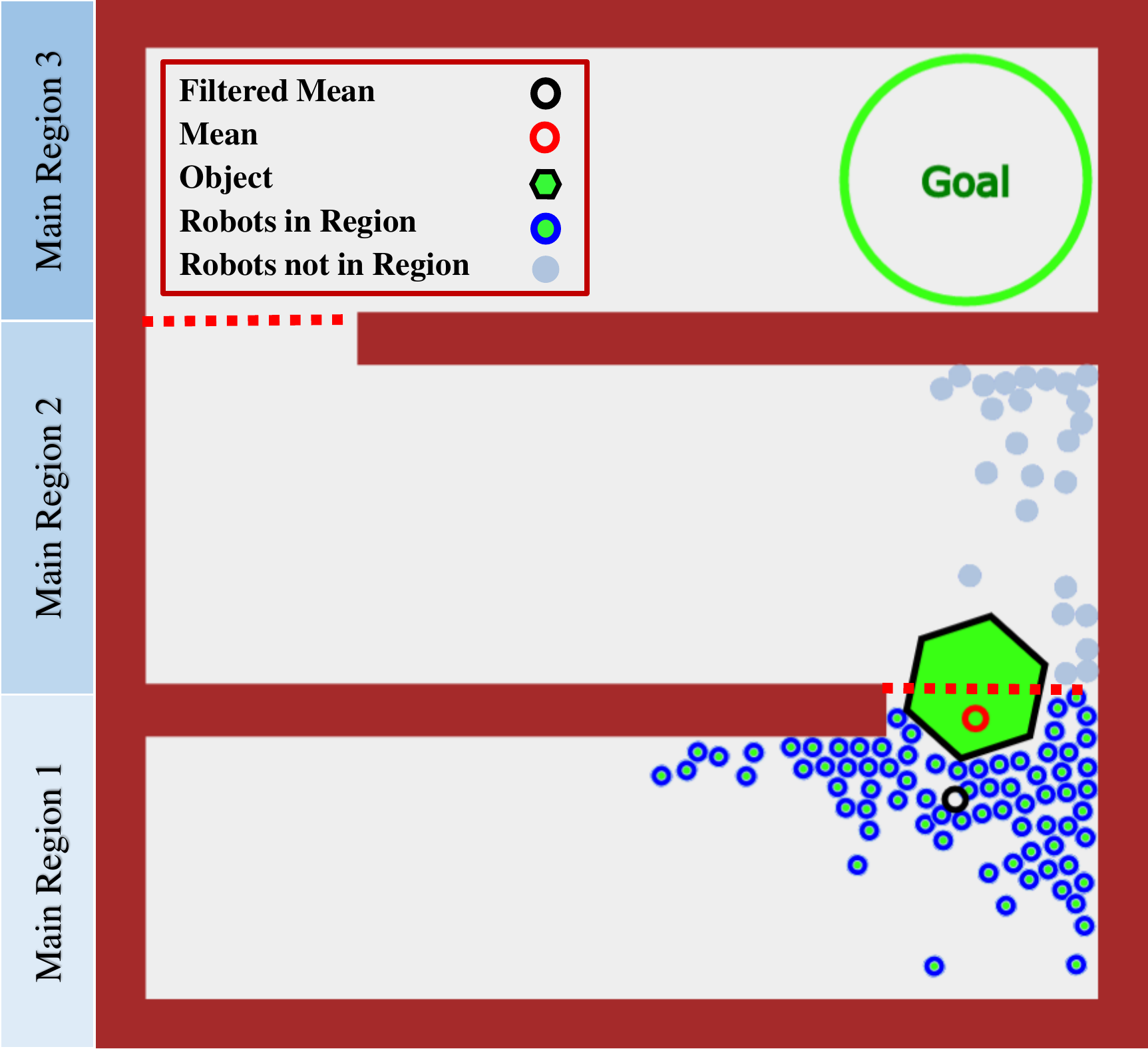} %,grid=true
	\put(-6,85){\textbf{b}}
	\end{overpic}
	~~
	\begin{overpic}[width=0.45\columnwidth]{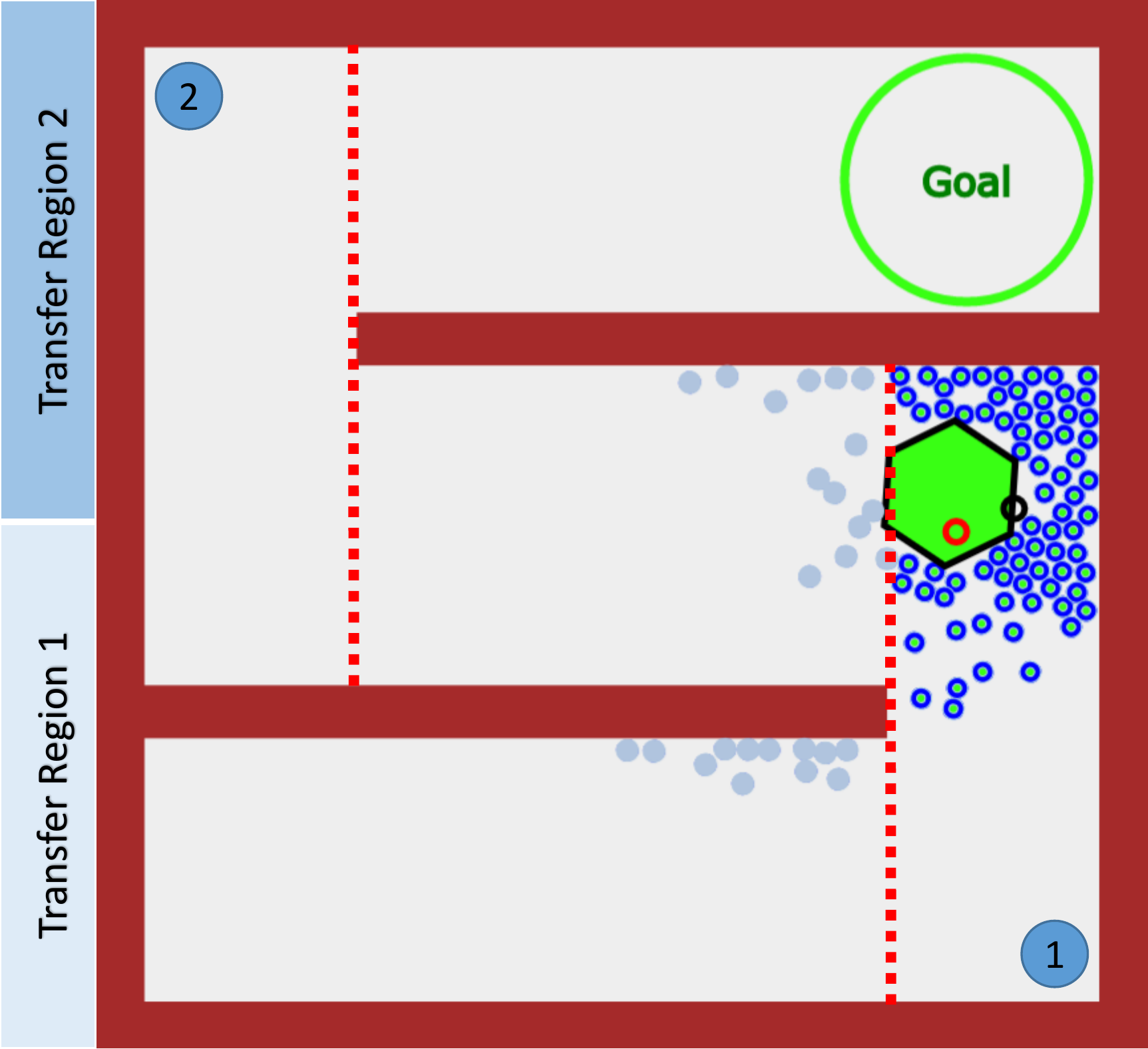}
	\put(-6,85){\textbf{c}}
	\end{overpic}
\end{center}
\vspace{-0.5em}
\caption{\label{fig:Region}  Outlier rejection state machine and regions.
}
\end{figure}

\subsection{Simulation results}\label{sec:AlgObjectManipulation}
 We use the hybrid hysteresis-based controller in Alg.~\ref{alg:MeanVarianceControl}  to track the desired position, while maintaining sufficient robot density to move the object by switching to minimize variance whenever variance exceeds a set limit:  $0.003 W$ and $0.006 W$ were added to the min and max variance limits from \eqref{eq:sigMaxMin}.
 The minimize variance control law \eqref{eq:PDcontrolVariance} is slightly modified to choose the nearest corner further from the goal than the object with an obstacle-free straight-line path to the object. 
The control algorithm  for object manipulation is listed in Alg.~\ref{alg:BlockPushing}. 

In rare cases during simulations the swarm may become trapped in a local minimum of \eqref{eq:potentialfield}.
If the swarm mean position does not change for five seconds, the swarm is assumed to be in a local minimum and is commanded to move toward the previous corner. As soon as the mean position changes, normal control resumes.

\begin{figure}
%  \vspace{-20pt}
  \begin{center}
\begin{overpic}[width=0.4\columnwidth]{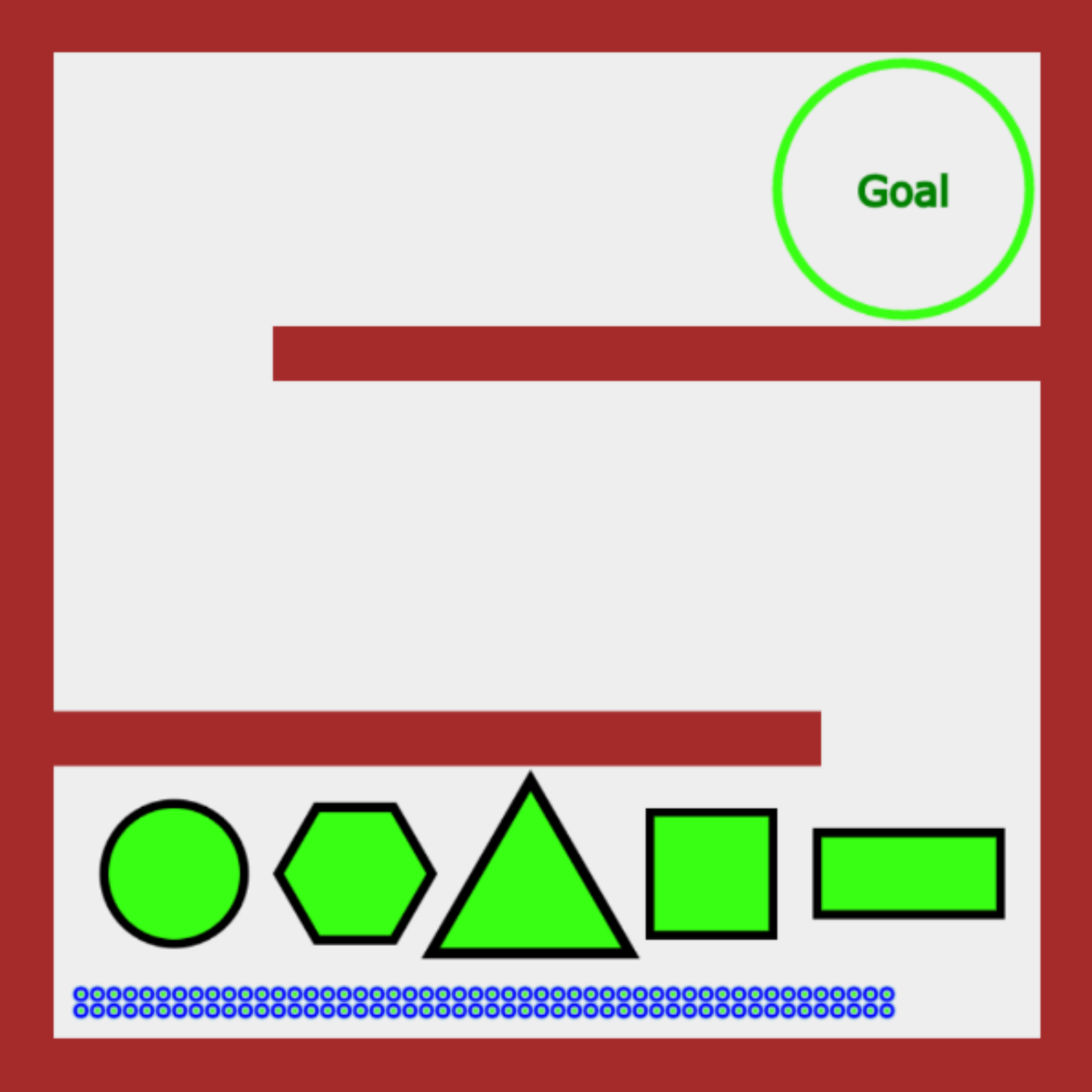}\end{overpic}
  \end{center}
%  \vspace{-1em}
\caption{\label{fig:Shapes} The six equal-area objects tested in simulation. %Six object shapes were tested.
\vspace{-1em}
}
\end{figure}

%\todo{add the timeout function: I added two lines.}

\begin{algorithm}
\caption{Object-manipulation controller for a robotic swarm.}\label{alg:BlockPushing}
\begin{algorithmic}[1]
\Require Knowledge of moveable object's center of mass $\mathbf{b}$; swarm mean $[\bar{x},\bar{y}]$ and variance $[\sigma_x^2, \sigma_y^2]$, each calculated using the regions function from \S \ref{subsec:OutlierRejection};  map of the environment
\State Compute optimal policy for object, according to \S \ref{subsec:objectpolicy}
\While{is not in goal region}
\State $\sigma^2 \gets \max{(\sigma_x,\sigma_y)}$
\If {$\sigma^2 > \sigma_{\rm{max}}^2$}
\While{$\sigma^2 > \sigma_{\rm{min}}^2$}
\State $ [x_{\rm{goal}}, y_{\rm{goal}}] \gets $ nearest corner in region
\State Apply \eqref{eq:PDcontrolPosition} to move toward $[x_{\rm{goal}}, y_{\rm{goal}}]$
\EndWhile
\Else  
%\If {$\mathrm{distance}( \mathbf{b}, [x_{goal}, y_{goal}] ) > k_1 r_b$}
%	\State$r_p \gets k_2 r_b$  \Comment{guarded move}
%	\Else
%	\State$r_p \gets k_3 r_b$  \Comment{pushing move}
%	\EndIf
%\State$r_p \gets k_3 r_b$  \Comment{pushing move}
\State Calculate $\mathbf{D}(\mathbf{b})$  \Comment{direction for object at $\mathbf{b}$}
\State Apply \eqref{eq:potentialfield}   \Comment{potential field for swarm}
\EndIf
\EndWhile
\end{algorithmic}
\end{algorithm}

%\subsection{Automated object manipulation (simulation)}
Fig.~\ref{fig:story} shows snapshots during an execution of this algorithm in simulation. 
Experimental results of parameters sweeps are summarized in Fig.~\ref{fig:AutoVeryParam}.  Each trial measured the time to deliver the object to the goal location.  The default parameter settings used 100 robots, a normalized weight of 1, a hexagon shape, and Brownian noise (applied once each simulation step) with $W=5$.  

\begin{figure}
\centering
%\renewcommand{\figwid}{0.19\columnwidth}
%\href{http://youtu.be/tCej-9e6-4o}{\begin{overpic}[width =\figwid]{story1.png}\put(6,15){T = 5 s}
%\end{overpic}
%\begin{overpic}[width =\figwid]{story2.png}\put(6,15){T = 12 s}
%\end{overpic}
%\begin{overpic}[width =\figwid]{story3.png}\put(6,15){T = 20 s}
%\end{overpic}
%\begin{overpic}[width =\figwid]{story4.png}\put(6,15){T = 25 s}
%\end{overpic}
%\begin{overpic}[width =\figwid]{story5.png}\put(6,15){T = 33 s}
%\end{overpic}}
\begin{overpic}[width =\columnwidth]{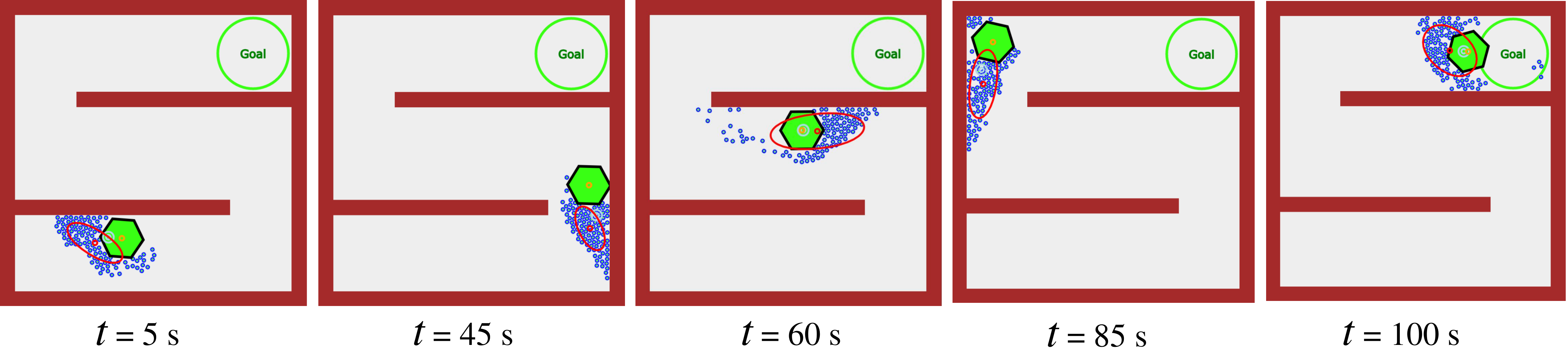}\end{overpic}
\caption{\label{fig:story}Snapshots showing an object manipulation simulation with 100 robots under automatic control (see also Extension 1).
}
\end{figure}

The interaction between the robots and object is impulsive so, like the study of impulsive pulling in  \cite{christensen2016let},  adding additional robots decreases completion time, but with diminishing returns. 
 After 75 robots, additional robots no longer can interact with the object and do not contribute to the task success. 
Brownian noise adds stochasticity.  This randomness can break the object free if it is stuck, but diminishes the effect of the control input.  
 Increasing noise increases completion time. 
 %Large amounts of noise caused failures, defined as trials lasting longer than 1000s.  With $W=50$, XX of 20 trials failed.  With $W=100$, XX of 20 trials failed.
The robots have limited force, so increasing the object weight increases completion time.  
Each shape was designed to have the same mass and area.
 Rectangles and squares tend to get stuck in the 90$^\circ$ workspace corners, and cause longer completion times than circles, triangles, and hexagons.

\begin{figure*}
\centering
\renewcommand{\figwid}{0.5\columnwidth}
\begin{overpic}[width =\figwid]{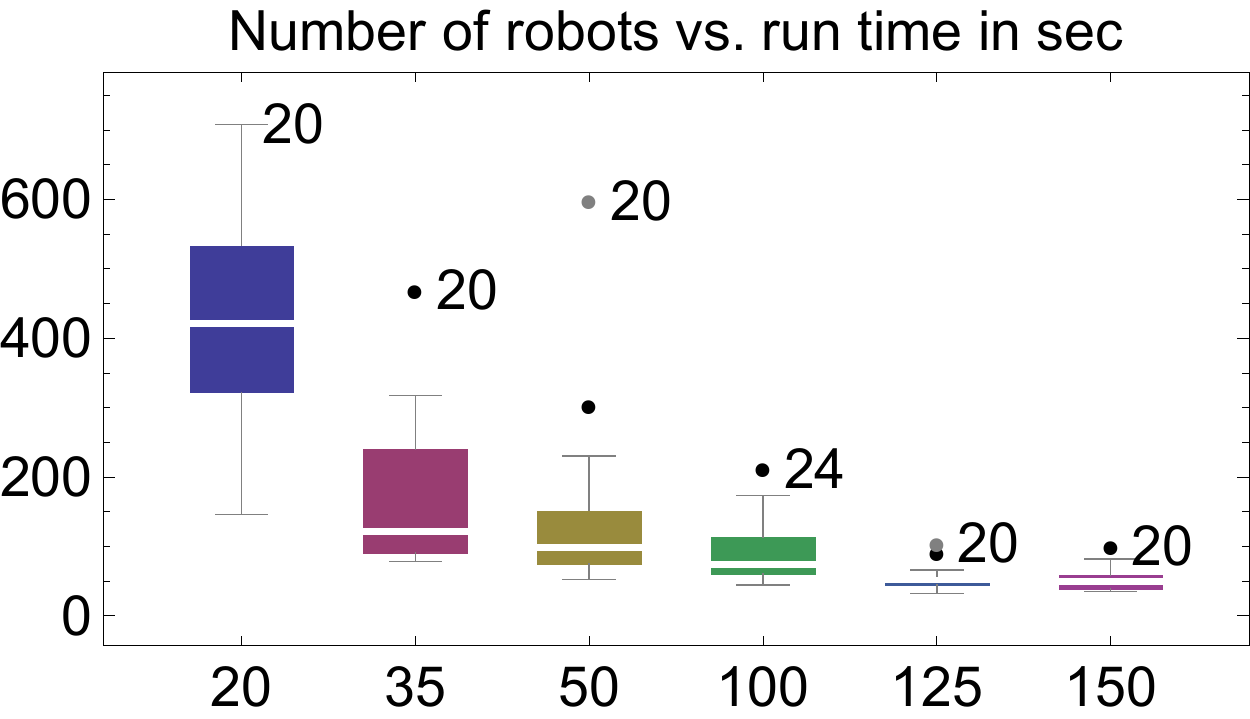}\put(1,55){a)}
\end{overpic}
\begin{overpic}[width =\figwid]{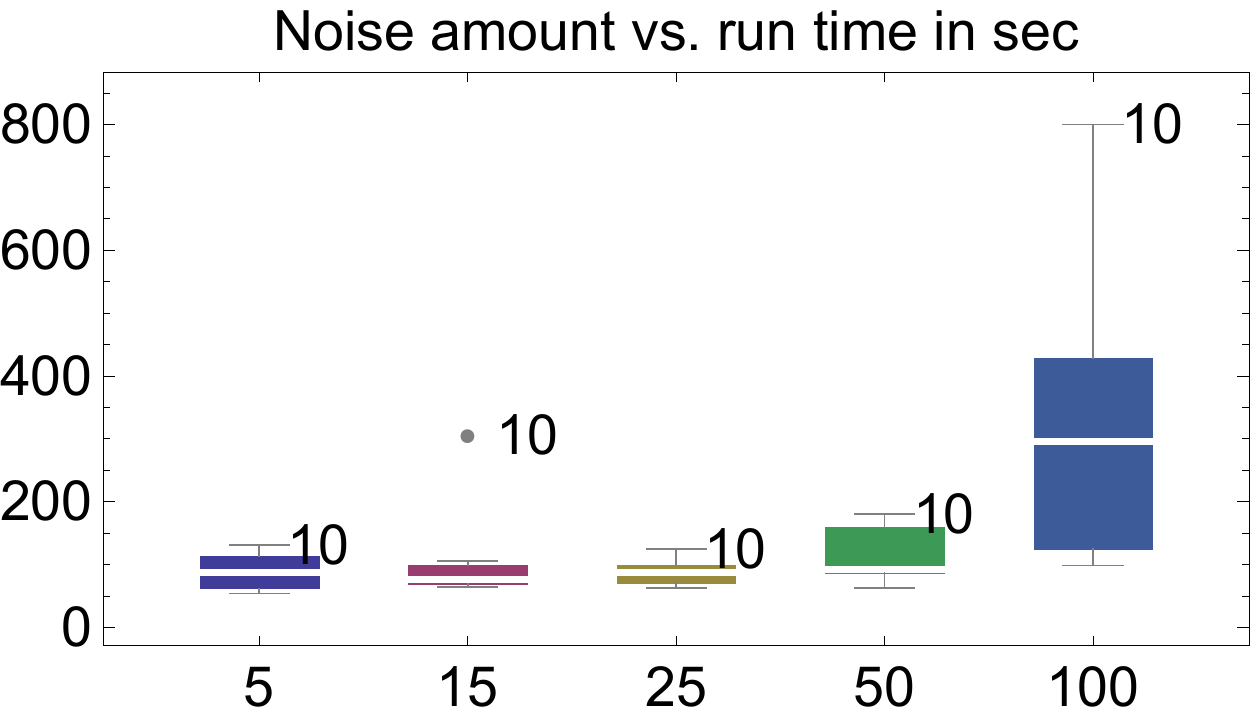}\put(1,55){b)}
\end{overpic}
\begin{overpic}[width =\figwid]{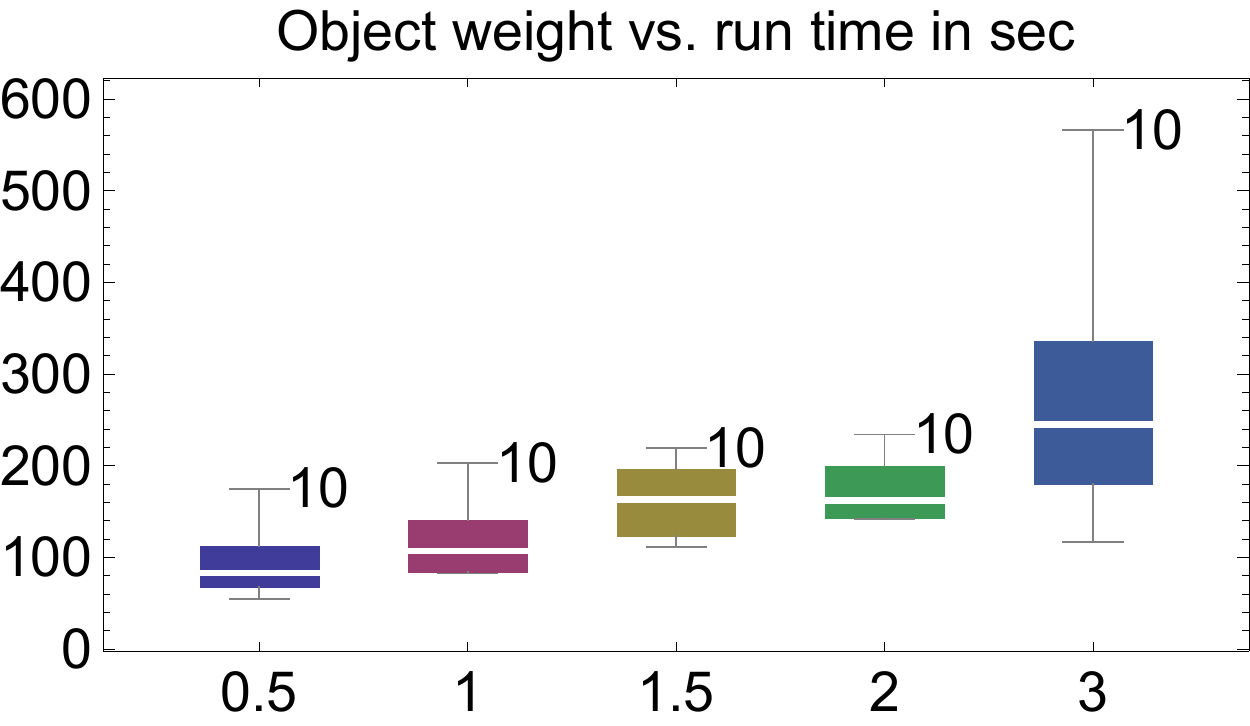}\put(1,55){c)}
\end{overpic}
\begin{overpic}[width =\figwid]{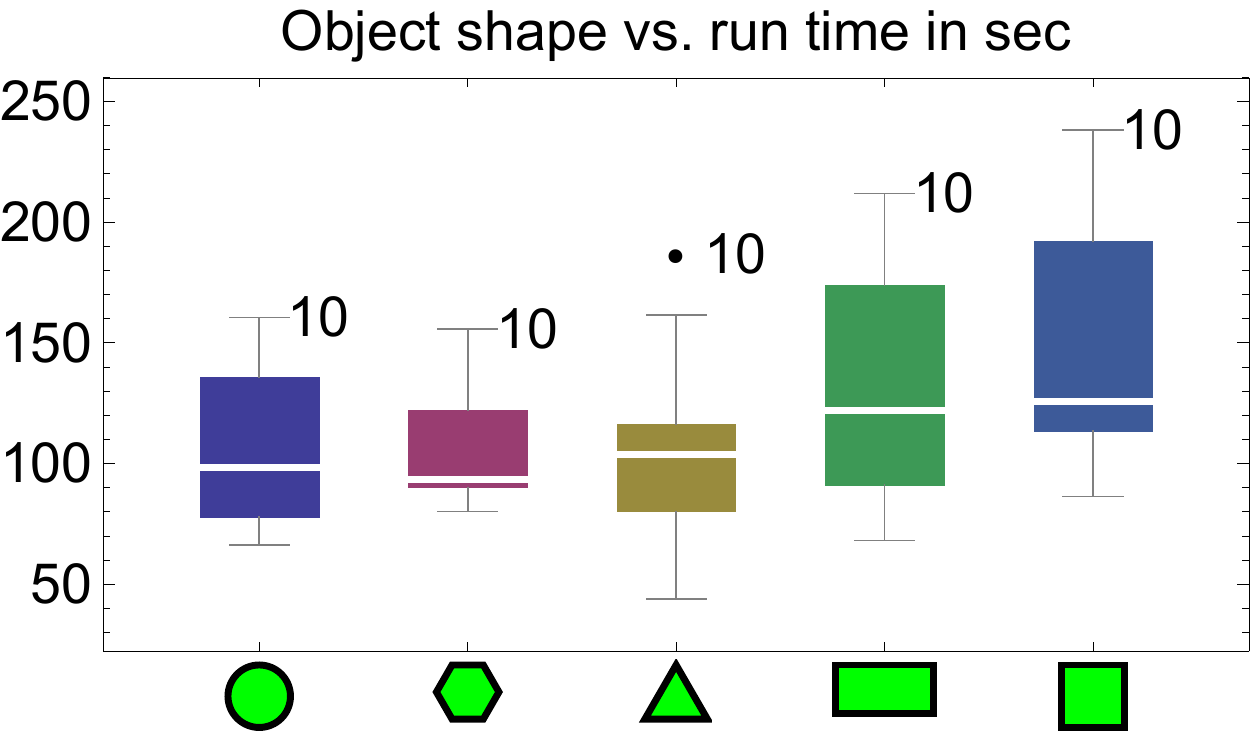}\put(1,55){d)}
\end{overpic}
\vspace{-0.5em}
\caption{\label{fig:AutoVeryParam}Parameter sweep simulation studies for a) number of robots, b) different noise values, c) object weight, and d) object shape.  Each bar is labelled with the number of trials. Completion time is in seconds.
%\vspace{-2em}
}
\end{figure*}

\section{Object manipulation with hardware robots}\label{sec:realExperiment}
%%%%%%%%%%%%%%%%%%%%%%%%%%%%%%%%%%%%%%%%%%%%%%%%%%%%%%%%%%%
Our experiments use centimeter-scale hardware systems called \emph{kilobots}.  While those are far larger than the micro scale devices we model, using kilobots allows us to emulate a variety of dynamics, while enabling a high degree of control over robot function, the environment, and data collection. The kilobot, reported in \cite{Rubenstein2012,rubenstein2014programmable}, is a low-cost robot designed for testing collective algorithms with large numbers of robots. It is available as an open-source platform or commercially from \cite{K-Team2015}.  Each robot is approximately 3 cm in diameter, 3 cm tall, and uses two vibration motors to move on a flat surface at speeds up to 1 cm/s.  Each robot has one ambient light sensor that is used to implement \emph{phototaxis},  moving towards a light source.

\subsection{Environmental setup}  
In these experiments as shown in Fig.~\ref{fig:setup}, we used $n$=100 kilobots, a 1.5 m$\times$1.2 m whiteboard as the workspace, and lights: four 50W LED floodlights  at the corners and four 30W LED floodlights on the sides of a 6 m square centered on the workspace and 1.5 m above the table. An Arduino Uno connected to an 8-relay shield controlled the lights.  

Above the table, an overhead machine vision system tracks the swarm. The vision system identifies obstacles by color segmentation, determines the corners  (used to decrease  variance), the object by color segmentation, and identifies robots using color segmentation and circle detection with a circular Hough transform. 

\begin{figure*}
\renewcommand{\figwid}{6cm}
\begin{center}
	\begin{overpic}[height=\figwid]{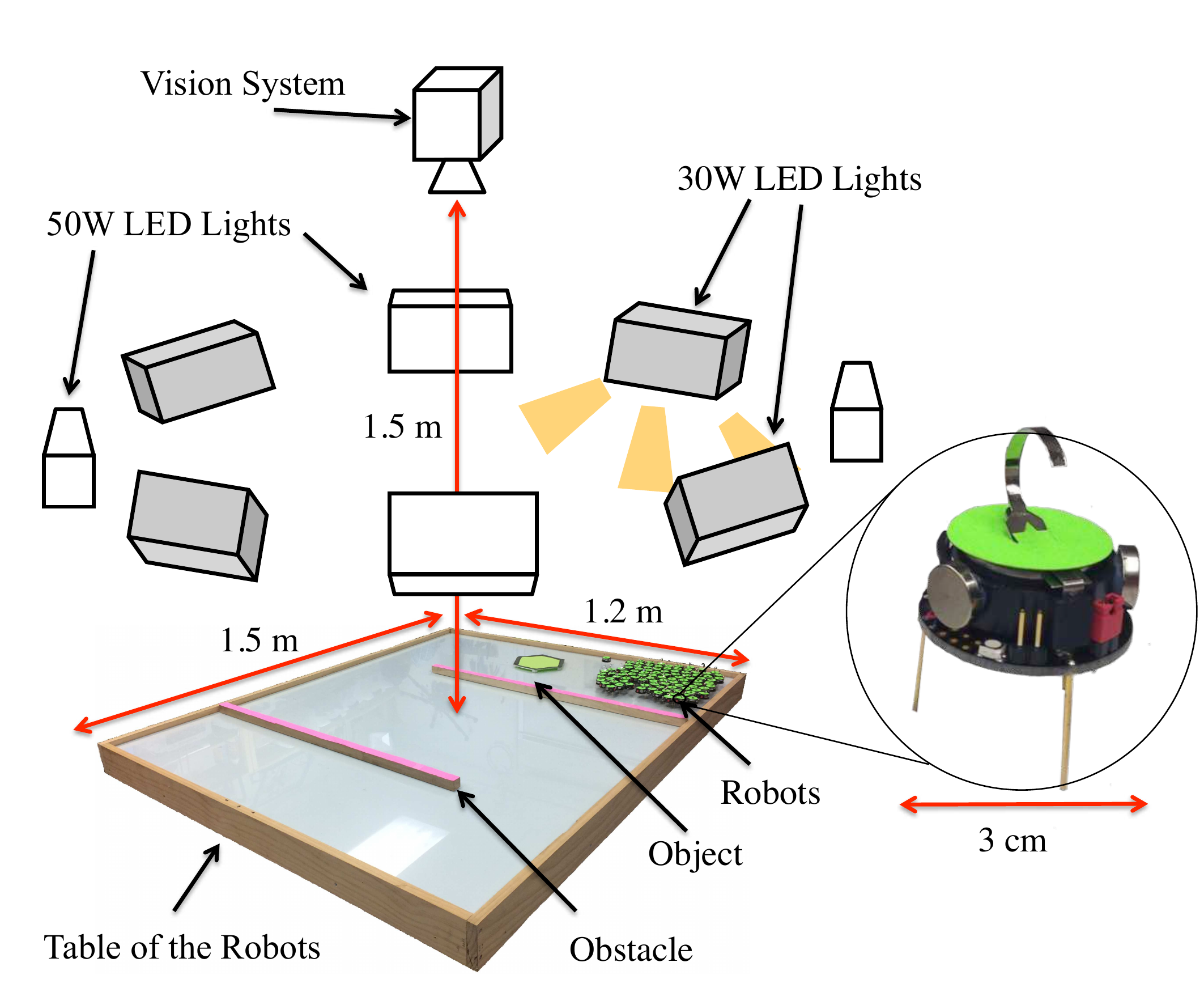}%\put(1,75){A}
	\end{overpic}\begin{overpic}[height=\figwid]{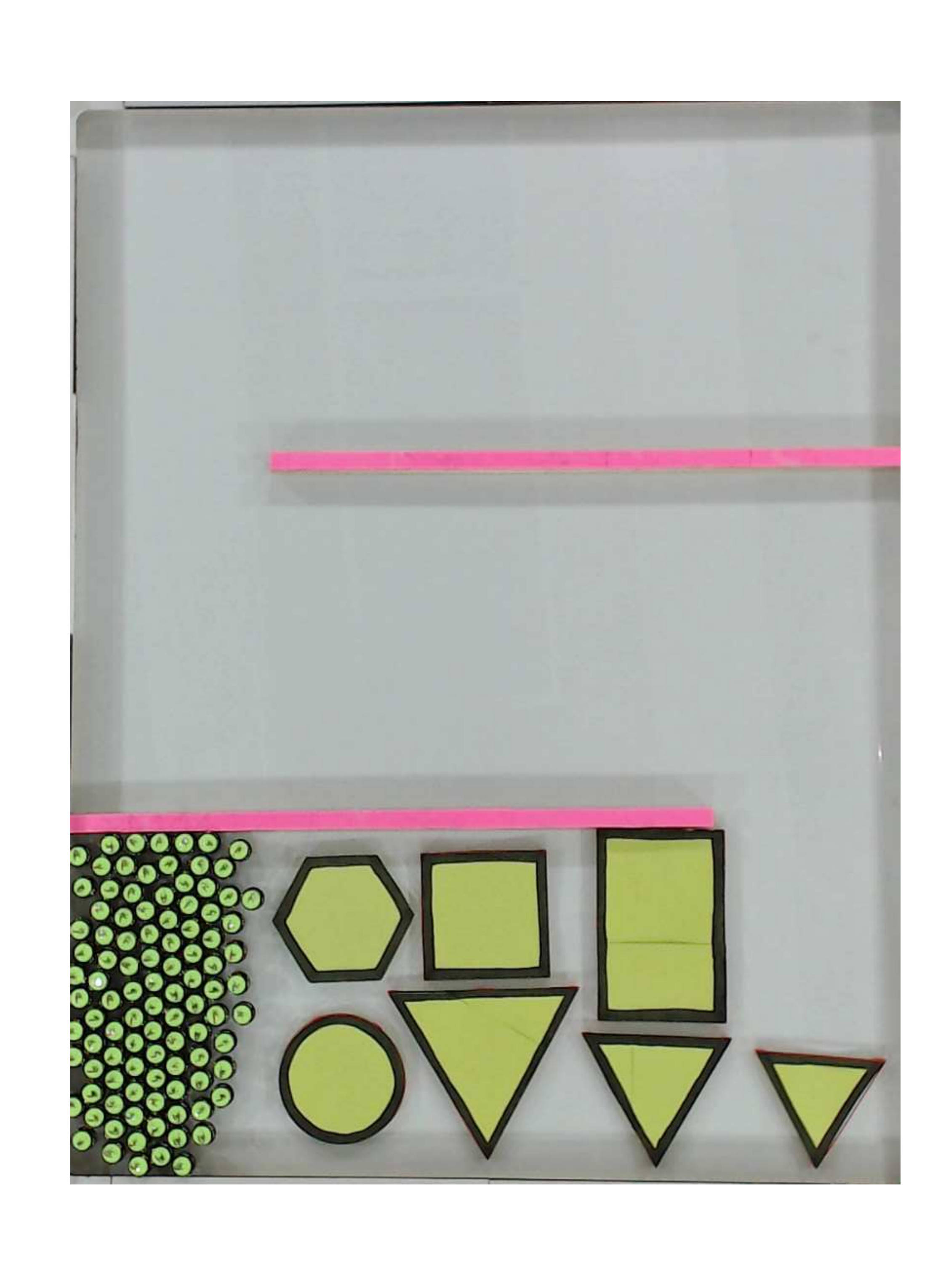}%\put(-0,85){B}
	\end{overpic}\begin{overpic}[height=\figwid]{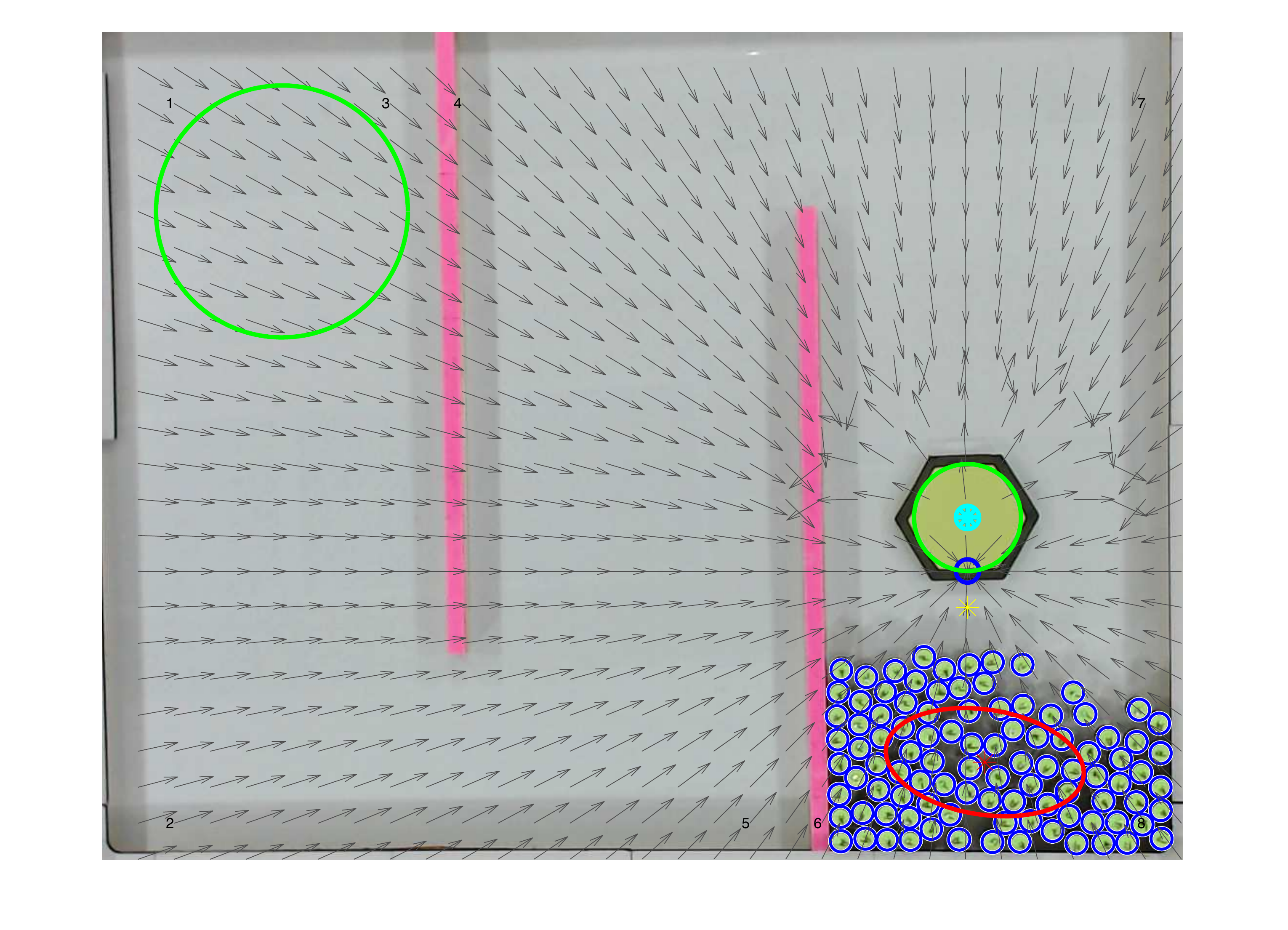}%\put(-0,85){C}
	\end{overpic}
\end{center}
\caption{\label{fig:setup}
Hardware platform. At right are the shapes used for hardware experiments and a visualization of the potential field. }
\end{figure*}

The objects were 3D printed from ABS plastic with a paper overlay. 
Shapes included a 325 cm$^2$ equilateral triangle, 
%Small Equilateral Triangle 140 cm$^2$,
 324 cm$^2$ square,
 281 cm$^2$ hexagon,
254 cm$^2$ circle, 
and a 486 cm$^2$ (18 cm$\times$27 cm)  rectangle, all shown in Fig.~\ref{fig:setup}. The laser-cut patterns for the neon green fiducial markers on the robots and 3D files for objects are available at our github repository,~\cite{Shahrokhi2016blocksimulations}.

\paragraph{Swarm mean control (hardware experiment)}

Unlike the PD controller \eqref{eq:PDcontrolPosition}, we cannot command a force input to the kilobots.  Instead, control is given by turning on one of eight lights.  The kilobots run a phototaxis routine where they search for an orientation that aligns them with the light source, and then move with an approximately constant velocity toward this light.   The kilobots oscillate along this orientation because they only have one light detector.  

We use the sign of \eqref{eq:PDcontrolPosition}, and choose the closest orientation to $\mathbf{D}(\mathbf{b})$ among the eight light sources.
Fig.~\ref{fig:realMean} shows that this limited, discretized control still enables regulating the mean position of a swarm of 100 robots.

\begin{figure}
\begin{center}
	\begin{overpic}[width=1.0\columnwidth]{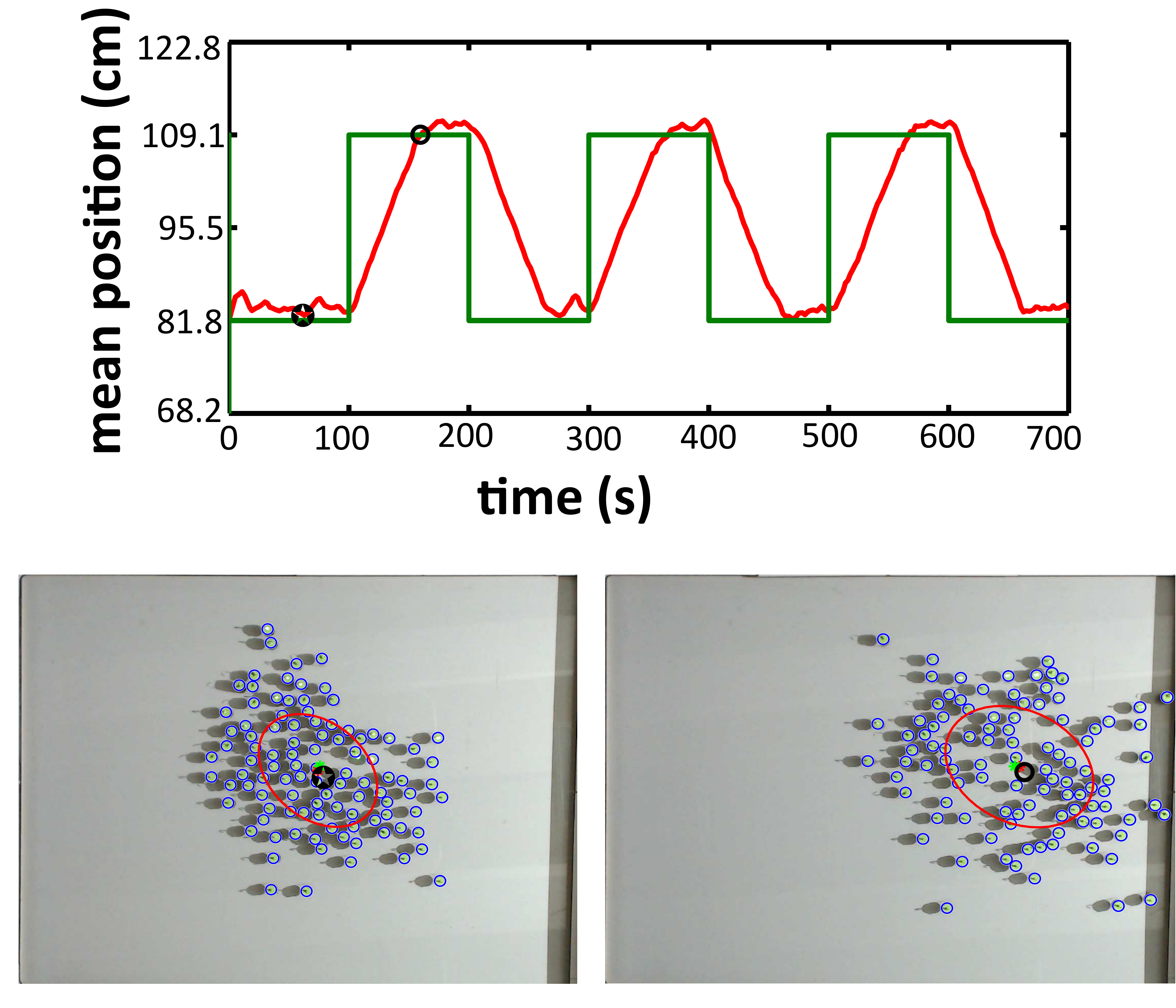}\put(6,30){\emph{t}= 90 s}\put(55,30){\emph{t}= 175 s}\end{overpic}
%	\begin{overpic}[width=0.35\columnwidth]{XYMeanControl.eps}\end{overpic}
\end{center}
\vspace{-1em}
\caption{\label{fig:realMean}
Regulating mean $x$ position of 100 kilobots using control law \eqref{eq:PDcontrolPosition}.
%Mean control plot with kilobots. %\todo{place a red star and a red circle at two spots on graph, put two images of the swar at right with a red star and red circle in left hand corner of these, and put a scale bar in image}
}
\end{figure}

\subsection{Automated object manipulation (hardware~experiment)}

The kilobots performed five successful runs manipulating a hexagonal object through an obstacle maze. Videos of these runs are in Extension 2. These hardware experiments represent the results of over 100 hours of trials.
Each trial used 100 kilobots. Trials two through five were performed in a row with no failures in between.  For each trial, fully charged kilobots were placed in the lower left-hand of the workspace, as shown in Fig.~\ref{fig:expSnapShot}.  The moveable object was placed in the lower center of the workspace.  {\sc Matlab} code for vision processing, the policy iteration of \S \ref{subsec:objectpolicy} and the algorithm of \S \ref{sec:AlgObjectManipulation} is available on {\sc Matlab} Central at \cite{Shahrokhi2015MDP}.
Trials were run until the object COM entered the goal region.  The trials ran for \{1465, 3457, 3000, 2162, 2707\} s.  This is $2558\pm771$ s (mean$\pm$std).

We also tested other object shapes. 
A circular object completed in 3155 s.    
A square object completed in 6871 s. 
A rectangle and three equilateral triangle objects of varying sizes failed in a total of nine runs. 
Manipulation failures occurred when the object was pushed into a corner, requiring torque to be unstuck.  
%This paper focuses on force, but not torque control. 
Swarm torque control is the subject of our ongoing research begun in \cite{Shahrokhi2016CASE}.

%Rules: http://www.ijrr.org/historic/MMVideoGuideLines.pdf

\begin{figure}
\centering
%\begin{overpic}[width=\columnwidth]{Snapshots.pdf}\put(6,29){\emph{t} = 3 s}\put(38,29){\emph{t} = 410 s}\put(70,29){\emph{t} = 710 s}
%\put(15,22){\emph{t} = 1374 s}\put(49,22){\emph{t} = 2185 s}\put(83,22){\emph{t} = 2703 s}
%\end{overpic}
\begin{overpic}[width=\columnwidth]{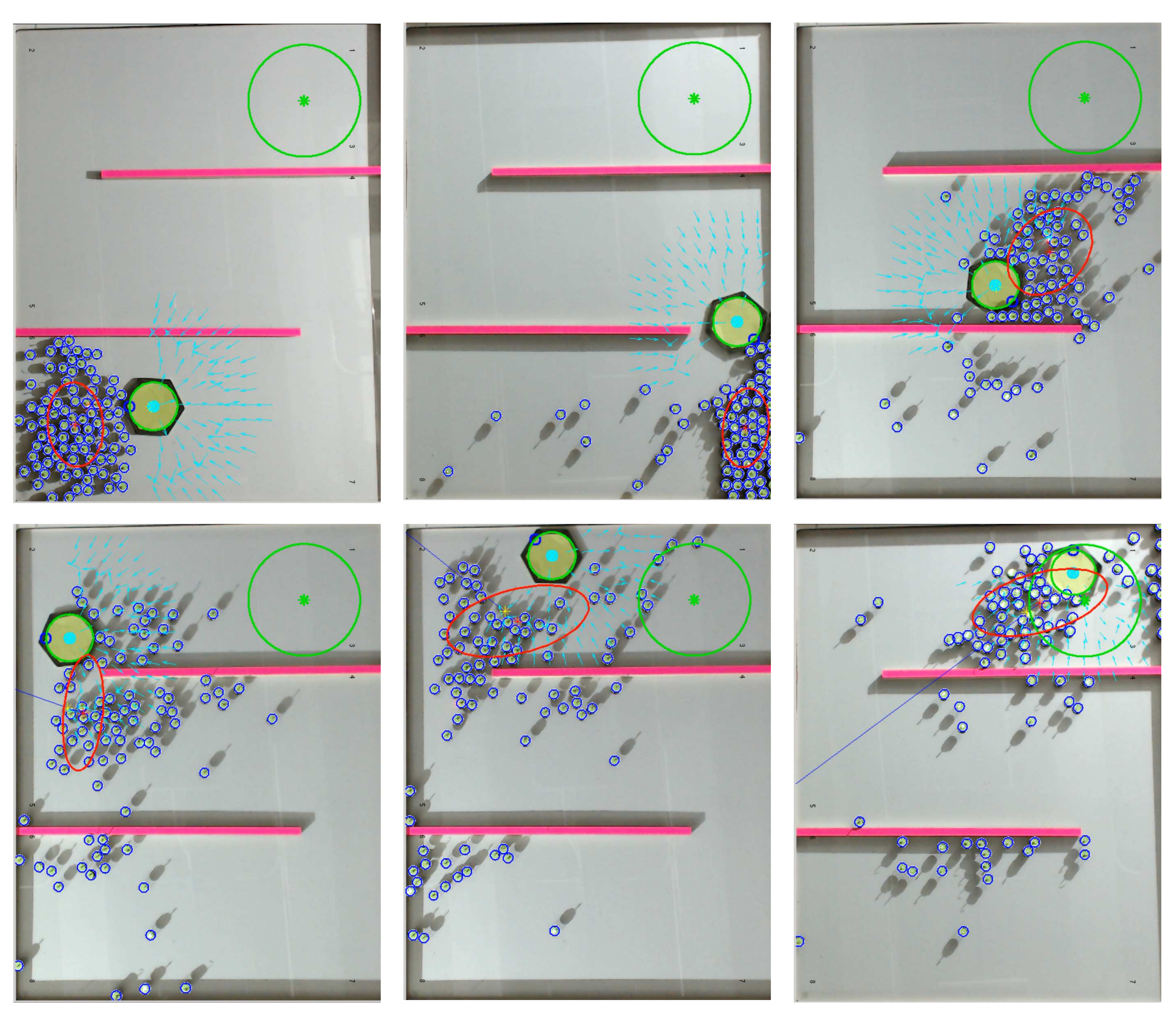}\put(6,80){\emph{t} = 3 s}\put(38,80){\emph{t} = 410 s}\put(70,80){\emph{t} = 710 s}
\put(15,8){\emph{t} = 1374 s}\put(48,8){\emph{t} = 2185 s}\put(80,8){\emph{t} = 2703 s}
\end{overpic}
%\vspace{-1em}
\caption{\label{fig:expSnapShot}Snapshots showing object manipulation experiment with 100 kilobots under automatic control. The automatic controller generates a policy to the goal, (see Extension 2).}
        \end{figure}
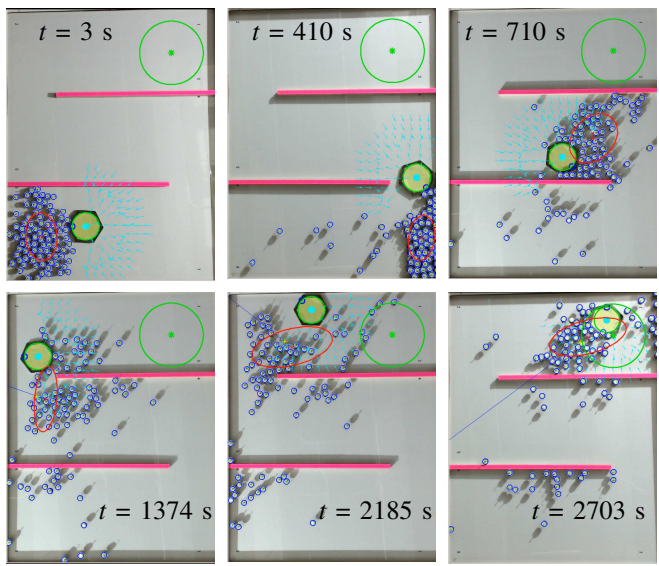

%%%%%%%%%%%%%%%%%%%%%%%%%%%%%%%%%%%%%%%%%%%%%%%%%%%%%%%%%%%
\section{Conclusion}\label{sec:conclusion}
%%%%%%%%%%%%%%%%%%%%%%%%%%%%%%%%%%%%%%%%%%%%%%%%%%%%%%%%%%%   
 
    %Micro- and nanorobotics have the potential to revolutionize many applications including targeted material delivery, assembly, and surgery.  
  The small size of micro and nano particles makes individual control and autonomy challenging, so currently these particles are steered by global control inputs such as magnetic fields or chemical gradients. To investigate this control challenge, this paper introduced \href{http://www.swarmcontrol.net}{SwarmControl.net}, an open-source tool for large-scale user experiments where human users steer swarms of robots to accomplish tasks.  Analysis of the game play results revealed benefits of measuring and controlling statistics of the swarm rather than full state feedback, robustness to IID noise, and small effects of varying population size of large swarms.

Inspired by the three lessons from \href{www.swarmcontrol.net}{swarmcontrol.net}, this paper designed controllers and controllability results using only the mean and variance of a particle swarm. 
We developed a hysteresis based controller to regulate the position and variance of a swarm. We designed a controller for object manipulation using policy iteration for path planning, regions for outlier rejection, and potential fields for minimizing moving the object backwards. 
All automatic controllers were implemented using 100 kilobots steered by the direction of a global light source.
These experiments culminated in an object manipulation task in a workspace with obstacles.
    
 % Manipulation by large populations of robots has many open questions. We invite other collaborators to submit their own experiments for large-scale trials to \href{http://www.swarmcontrol.net}{SwarmControl.net}.

Our future goal is to perform assembly using particle swarms to manipulate and attach components. This task requires controlling the position and orientation of components, manipulating them through obstacles and other components, and applying force and torque to components. This work provides foundational algorithms and techniques for steering swarms, object manipulation, and addressing obstacle fields, but there are many opportunities to extend the work.

Topics of interest include control with nonuniform flow such as fluid in an artery, gradient control fields like that of an MRI, competitive playing, multi-modal control, optimal-control, and targeted drug delivery in a vascular network.

\section*{Acknowledgements}\label{sec:Acknowledgements}
We thank our anonymous reviewers and Dylan Shell for their substantive comments which improved the presentation and content of this paper.
\balance

%\begin{dci}
%The Authors declare that there is no conflict of interest.
%\end{dci}
%\begin{funding}
%%RULES: http://www.nsf.gov/pubs/policydocs/pappguide/nsf16001/aag_6.jsp
%This work was supported by the National Science Foundation under Grant No.\ 
%\href{http://nsf.gov/awardsearch/showAward?AWD_ID=1553063}{ [IIS-1553063]}.
%\end{funding}
%\newpage
%\begin{sm}
%\begin{itemize}
%\item Extension 1:  simulation video of the parameter sweeps in Fig.~\ref{fig:AutoVeryParam} with robot swarms manipulating objects.
%\item Extension 2: hardware experiment videos of object manipulation with a swarm of 100 kilobots.
%\end{itemize}
%\end{sm}

%\input{11-MultiMediaAppendix.tex}
   
\bibliographystyle{IEEEtran}
\bibliography{IEEEabrv,SwarmControlJournal}

\end{document}